\documentclass{article}

\PassOptionsToPackage{numbers, sort, compress}{natbib}
\usepackage[final]{neurips_2025}

\usepackage{microtype}
\usepackage{float}
\usepackage{graphicx}
\usepackage{subcaption}
\usepackage{booktabs}
\usepackage{wrapfig}
\usepackage{amsmath}
\usepackage{amssymb}
\usepackage{mathtools}
\usepackage{amsthm}

\usepackage[pagebackref=true]{hyperref}

\usepackage{Shorthands}

\usepackage{amssymb}
\usepackage{amsmath}
\usepackage{latexsym}
\usepackage{graphicx}
\usepackage{color}
\usepackage{url}
\usepackage{Shorthands}
\usepackage{listings}
\usepackage{courier}
\usepackage{comment}
\usepackage{algorithm}
\usepackage{hyperref}
\renewcommand{\hat}{\widehat}

\renewcommand{\P}{{\mathbf P}}

\renewcommand{\P}{\ensuremath{\mathbf{P}}}

\usepackage{mathtools}

\renewcommand{\qed}{\hfill\blacksquare}
\DeclareMathOperator{\train}{train}
\DeclareMathOperator{\calib}{cal}

\newcommand{\Yhat}{\ensuremath{\widehat{Y}}}
\theoremstyle{plain}
\newtheorem{theorem}{Theorem}[section]

\newtheorem{corollary}[theorem]{Corollary}
\theoremstyle{definition}
\newtheorem{definition}[theorem]{Definition}

\theoremstyle{remark}

\newtheorem{problem}[theorem]{Problem}
\newtheoremstyle{boldremark}
    {\dimexpr\topsep/2\relax} % space above
    {\dimexpr\topsep/2\relax} % space below
    {}          % body font
    {}          % indent amount
    {\bfseries} % theorem head font
    {.}         % punctuation after theorem head
    {.5em}      % space after theorem head
    {}          % theorem hed spec. (empty = "normal")

\theoremstyle{boldremark}
\newtheorem{brem}[theorem]{Remark} % remarks are numbered within sections

\title{Uncertainty-Calibrated Prediction of Randomly-Timed Biomarker Trajectories with Conformal Bands}

\begin{document}

\author{
\textbf{Vasiliki Tassopoulou}$^{1,*}$ \quad
\textbf{Charis Stamouli}$^{2,*}$ \quad \\
\textbf{Haochang Shou}$^{3}$ \quad
\textbf{George J. Pappas}$^{2}$ \quad
\textbf{Christos Davatzikos}$^{1}$
}

\maketitle

\begin{center}
$^{1}$ Center for AI and Data Science for Integrated Diagnostics (AI2D), Perelman School of Medicine, University of Pennsylvania, Philadelphia, PA, USA \\
$^{2}$Department of Electrical and Systems Engineering, University of Pennsylvania, PA, USA \\
$^{3}$Department of Biostatistics and Epidemiology, University of Pennsylvania, PA, USA \\
E-mail: vtass@seas.upenn.edu, stamouli@seas.upenn.edu, hshou@pennmedicine.upenn.edu,\\
pappasg@seas.upenn.edu, christos.davatzikos@pennmedicine.upenn.edu
\end{center}

\footnotetext[1]{Equal contribution}

\begin{abstract}
Despite recent progress in predicting biomarker trajectories from real clinical data, uncertainty in the predictions poses high-stakes risks (e.g., misdiagnosis) that limit their clinical deployment. To enable safe and reliable use of such predictions in healthcare, we introduce a conformal method for uncertainty-calibrated prediction of biomarker trajectories resulting from randomly-timed clinical visits of patients. Our approach extends conformal prediction to the setting of randomly-timed trajectories via a novel nonconformity score that produces prediction bands guaranteed to cover the unknown biomarker trajectories with a user-prescribed probability. We apply our method across a wide range of standard and state-of-the-art predictors for two well-established brain biomarkers of Alzheimer's disease, using neuroimaging data from real clinical studies. We observe that our conformal prediction bands consistently achieve the desired coverage, while also being tighter than baseline prediction bands. To further account for population heterogeneity, we develop group-conditional conformal bands and test their coverage guarantees across various demographic and clinically relevant subpopulations. Moreover, we demonstrate the clinical utility of our conformal bands in identifying subjects at high risk of progression to Alzheimer’s disease. Specifically, we introduce an uncertainty-calibrated risk score that enables the identification of 17.5\% more high-risk subjects compared to standard risk scores, highlighting the value of uncertainty calibration in real-world clinical decision making. Our code is available at \href{https://github.com/vatass/ConformalBiomarkerTrajectories}{\texttt{github.com/vatass/ConformalBiomarkerTrajectories}}.
\end{abstract}

\vspace{-0.3cm}
\section{Introduction}\label{sec:Introduction}
\vspace{-0.3cm}

Predicting biomarker trajectories is crucial for monitoring disease progression and improving patient outcomes through early intervention. For example, in neurodegenerative diseases such as Alzheimer’s, volumetric biomarkers—most notably hippocampal atrophy—are closely associated with clinical decline and are widely used to track disease progression  \cite{Jones2010,Maheux2023Forecasting,Vecchio2025}. Given the clinical importance of biomarker trajectories, recent advances in machine learning have led to a growing number of models designed to predict biomarker evolution from longitudinal data \cite{schiratti:hal-01540367, NGUYEN2020117203, NEURIPS2021_c7b90b0f,marinescu2021alzheimersdiseasepredictionlongitudinal,tassopoulouadaptive}. However, inter-individual variability in disease progression and measurement noise inherent in imaging data make predictions inherently uncertain, with predictors not being able to determine exact future values with complete confidence. If uncertainty is not properly accounted for, predictions can cause high-risk subjects to be misclassified as stable. Such potential misclassifications hinder the deployment of biomarker predictors in clinical workflows, where failure to identify high-risk individuals can delay urgently needed intervention or exclude such individuals from clinical trials. To enable safe and reliable decision making in healthcare, uncertainty calibration of the learned predictors is important.

% However, inter-individual variability in disease progression and measurement noise inherent in imaging data make predictions inherently uncertain—models cannot determine exact future values with complete confidence. Without properly quantifying this predictive uncertainty, clinicians lack crucial information about the reliability of individual predictions. For instance, a model might predict mild decline for a patient while having low confidence in this prediction—the true trajectory could range from stability to severe decline. Failing to communicate this uncertainty could lead to high-risk subjects being misclassified as stable, potentially delaying critical interventions

A popular approach for equipping predictors with reliable uncertainty estimates is given by conformal prediction \cite{Vovk2005,Shafer2008,Papadopoulos2008}. Conformal prediction is highly versatile, accommodating any black-box predictive model (e.g., neural networks) and any data distribution (e.g., non-Gaussian). The first step in the approach is to evaluate the learned model's prediction error, as characterized by a so-called \textit{nonconformity score}, on a held-out calibration dataset. Assuming that the calibration and test data are sampled independently from the same probability distribution\footnote{Conformal methods typically require that the calibration and test data are \textit{exchangeable}---a milder condition than that of independent and identically distributed data (see Appendix~\ref{sec:Exchangeability}).}, the prediction error on the calibration data can be used to estimate the prediction error on the test data. Thus, model predictions can be turned into regions that are guaranteed to cover the true observations with arbitrarily high probability. 

\begin{figure}%{r}{0.53\columnwidth}
\vspace{-0.2cm} % tune to align with paragraph
\centering
\includegraphics[width=0.5\linewidth]{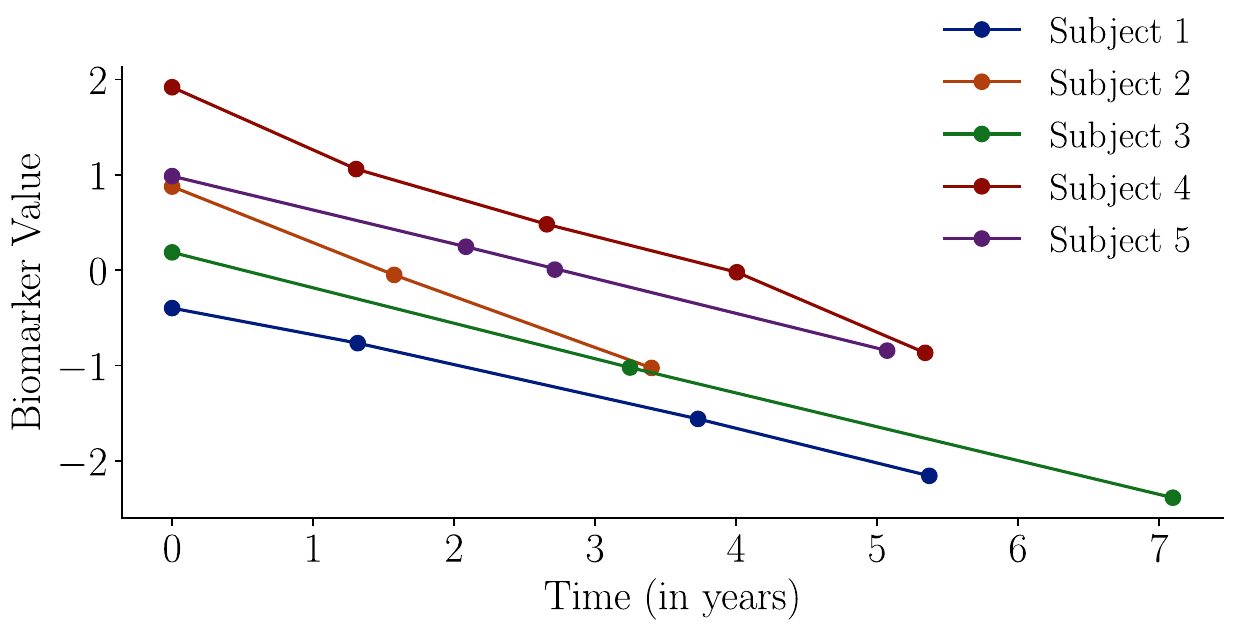}
\caption{Illustration of randomly-timed biomarker trajectories from five subjects. Time is measured with respect to the first biomarker observation of each subject. Note the varying trajectory length and the different time points of observations across subjects.}
\label{fig:irregular_patient_data}
\vspace{-0.4cm}
\end{figure}

In the context of trajectory prediction, existing conformal methods focus on fixed-time trajectories \cite{Gibbs2021,Stankeviciute2021,Xu2021,Lekeufack2024,Stamouli2024}. However, biomarker datasets typically contain randomly-timed biomarker trajectories due to missed visits and variable scheduling in clinical studies (see Figure~\ref{fig:irregular_patient_data}). Such trajectories create a more complex setting for designing conformal prediction bands with guaranteed coverage properties, which are critical for safely delivering early diagnoses and effectively guiding clinical interventions.

In this paper, we propose a novel conformal method for uncertainty-calibrated prediction of randomly-timed biomarker trajectories. Our approach derives prediction bands that are guaranteed to cover the unknown biomarker trajectories with a user-prescribed probability based on a new nonconformity score. We test our method on a wide range of standard and state-of-the-art predictors for trajectories of hippocampal and ventricular volume—two well-established brain biomarkers used to monitor the onset and progression of Alzheimer’s disease. To further capture variability across clinically meaningful subpopulations, we design group-conditional conformal bands that ensure the desired coverage within each subpopulation of interest. In our case study, we derive group-specific prediction bands for five distinct population stratifications based on sex, race, diagnosis, education level, and APOE4 allele status---a genetic risk factor related to Alzheimer's disease. These five covariates are selected to ensure equitable guarantees for subpopulations that are underrepresented in clinical studies or have significant risk of progressing to Alzheimer's disease. Next, we leverage our conformal prediction bands in a downstream clinical decision task, where the goal is to identify Mildly Cognitively Impaired (MCI) patients at high risk of progression to Alzheimer’s disease. In particular, we introduce an uncertainty-calibrated version of a standard risk score employed for this purpose, which is empirically observed to enable more inclusive and safety-aware subject selection in scenarios such as clinical trial enrichment or early clinical intervention planning.

\begin{figure}%{r}{0.53\columnwidth}
\vspace{-0.1cm} % tune to align with paragraph
\centering
\includegraphics[width=\linewidth]{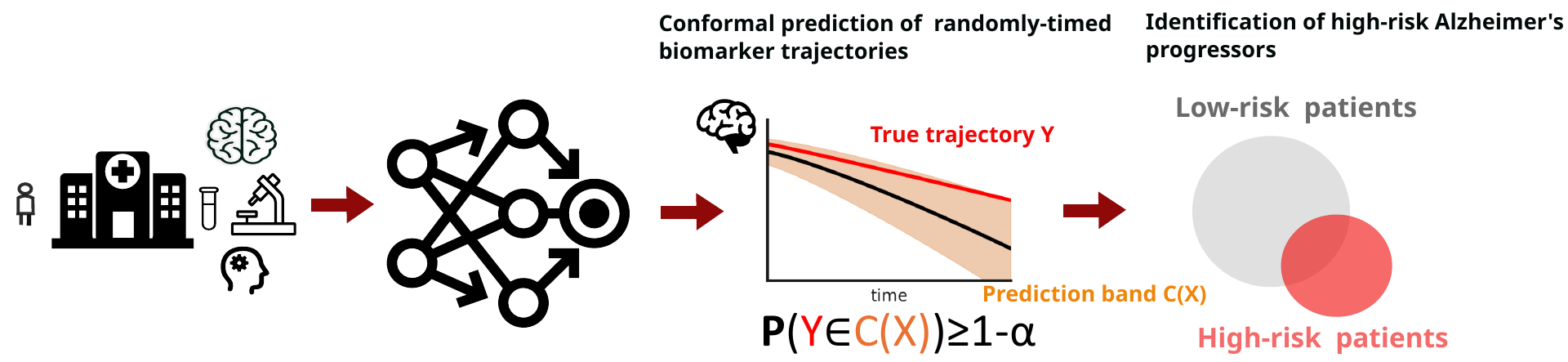}
\caption{We predict randomly-timed biomarker trajectories using arbitrary predictors (e.g., neural networks) and bound the prediction uncertainty within high-confidence prediction bands (orange band).  Using these bands, we develop an uncertainty-calibrated method of identifying high-risk patients.}
\label{fig:method_summary}
\vspace{-0.4cm}
\end{figure}

Our contributions can be summarized as follows:
\begin{itemize}
\item We introduce a new conformal prediction framework tailored to randomly-timed trajectories. Our approach leverages a nonconformity score defined over multiple time points, similar to \citet{Yu2023,cleaveland2024time}, with the important distinction that the time points in our setting are not fixed but random, reflecting real-world clinical visits of subjects.

\item We test our approach on two well-established neurodegenerative biomarkers—hippocampal volume and ventricular volume—using neuroimaging data from real clinical studies. By conformalizing a wide range of standard and state-of-the-art predictors, we demonstrate that our method achieves nominal coverage while maintaining tight prediction bands, outperforming baseline uncertainty-calibration methods that rely on simplifying assumptions for the predictor and the data distribution.

\item To address population heterogeneity, we stratify our calibration data into groups and design group-conditional conformal prediction bands with coverage guarantees within each group. We test these guarantees across five demographic and clinically relevant population stratifications based on sex, race, diagnosis, education level, and a genetic risk factor related to Alzheimer's disease.

\item We demonstrate the clinical utility of our uncertainty-calibrated conformal prediction bands in a downstream clinical task, where the goal is to identify MCI subjects at high risk of progression to Alzheimer’s disease. To this end, we introduce a novel uncertainty-calibrated risk score which leverages our conformal bands to provide a high-confidence worst-case bound for the biomarker's rate of change over time. This score enables the \textbf{identification of 17.5\% more high-risk MCI subjects} who progress to Alzheimer’s disease compared to standard scores.
\end{itemize}

\section{Related Work}\label{sec:related_work}
\textbf{Uncertainty-Calibrated Prediction of Biomarker Trajectories.}
Uncertainty-calibrated predictors for biomarker trajectories integrate uncertainty quantification into their architecture. Bayesian methods achieve this by employing probabilistic frameworks to model uncertainty. For instance, Gaussian Processes (GPs) \citep{Rasmussen2004} utilize kernel-defined priors over functions to encode structural assumptions, such as smoothness. Their posterior distributions quantify prediction uncertainty through covariance-based inference under Gaussian likelihoods. Several studies \citep{chung2019deepmixedeffectmodel, Lorenzi_2019, pmlr-v106-rudovic19a, ABINADER2020116266, pmlr-v193-tassopoulou22a, tassopoulouadaptive} involve predictors with inherent uncertainty measures based on Gaussian distribution assumptions. In contrast, \citet{Lin2022} introduce a conformal variant for fixed-time biomarker trajectories, ensuring asymptotic longitudinal coverage without any distributional assumptions.
% Additionally, \citet{mobeirek2024uncertaintyquantificationalzheimersdisease} evaluated four uncertainty quantification techniques—including Monte Carlo dropout, variational inference, and ensembles—for predicting cognitive scores.

\textbf{Conformal Prediction for Fixed-Time Trajectories.}
Conformal methods for predicting fixed-time trajectories have been developed for settings with single time series data \cite{Xu2023,Angelopoulos2024,Gibbs2024} and multiple time series data \cite{Stankeviciute2021,Lin2022,Sun2022,Lindemann2023a,Yu2023,cleaveland2024time}. Our setting differs from the above, involving data with \textit{generally non-temporal} inputs (e.g., sex) and \textit{randomly-timed} trajectory outputs (e.g., hippocampal-volume measurements taken at random time points for each subject). We focus on the latter line of works, whose setting is a special case of ours.

\citet{Lin2022} present a conformal prediction method with asymptotic longitudinal coverage guarantees. \citet{Stankeviciute2021} and \citet{Lindemann2023a} design conformal prediction bands with finite-sample coverage guarantees, by leveraging a union bound argument. \citet{Sun2022} propose using copulas to model the uncertainty of predictions at multiple time points. Their approach ensures coverage with narrower prediction bands, but is limited to situations with ample calibration data, as noted by the authors. \citet{Yu2023} and \citet{cleaveland2024time} overcome this limitation by defining a normalized nonconformity score jointly over multiple time points. Their framework provides narrow prediction bands with simultaneous coverage guarantees over the entire trajectory length.  In the next section, we extend their approach to our setting of randomly-timed trajectories, where observations are collected at varying time points.

\vspace{-0.3cm}
\section{Conformal Prediction for Randomly-Timed Trajectories}\label{sec:Conformal_Prediction_for_Irregular_Time_Series} 
\vspace{-0.3cm}

In this section, we introduce a novel conformal prediction method tailored to settings with randomly-timed trajectories. In Subsection~\ref{subsec:Problem_Formulation}, we present the problem of conformal prediction for data with vector-valued inputs and randomly-timed trajectory outputs. In Subsection~\ref{subsec:Conformal_Prediction Bands_for_Irregular_Trajectories}, we derive conformal prediction bands with coverage guarantees for our setting, inspired by the approach of \citet{Yu2023} and \citet{cleaveland2024time} for fixed-time trajectories. 

\vspace{-0.3cm}
\subsection{Problem Formulation}\label{subsec:Problem_Formulation}
\vspace{-0.3cm}

Our setting of randomly-timed trajectories is characterized by a triplet of random variables $(X,\calT,Y)$, where $X\in\setR^d$ denotes the input, $\calT\subseteq\setN_+$ the set of time points at which observations $Y_t\in\setR$ are collected, and $Y:=\{Y_t:t\in\calT\}$ the corresponding trajectory output. In the context of biomarkers, the triplet $(X,\calT,Y)$ can be interpreted, for instance, in the following way: i) the input $X$ may involve demographic covariates (e.g., sex, race) as well as the biomarker observation on the patient's first clinical visit, ii) the set $\calT$ may include the times of the patient's subsequent visits, and iii) the output $Y$ the corresponding biomarker observations at these times. We note the implicit dependence of the trajectory output $Y$ on the set of time points $\calT$.

Suppose we have a dataset $D:=\{(X^{(i)},\calT^{(i)},Y^{(i)})\}_{i=1}^N$ such that for each test example $(X,\calT,Y)$, the random variables $(X^{(1)},\calT^{(1)},Y^{(1)}),\ldots,(X^{(N)},\calT^{(N)},Y^{(N)})$, and $(X,\calT,Y)$  are \textit{exchangeable}\footnote{Exchangeability implies that the joint probability distribution of a sequence of random variables remains unchanged under any permutation of the variables (see definition in Appendix~\ref{sec:Exchangeability}).\label{fn:exchangeability2}}. We note that exchangeability is weaker than the standard assumption of independent and identically distributed data, imposing a reasonable condition for biomarker datasets, where each triplet $(X^{(i)},\calT^{(i)},Y^{(i)})$ corresponds to a different subject. Our goal is to leverage the dataset $D$ and the input $X$ to design a high-confidence prediction band for the unknown trajectory $Y$ (see Figure~\ref{fig:example_bands}). Our problem is formalized in the following statement. 

\begin{problem}
Consider a given dataset $D$, consisting of $N$ data $(X^{(1)},\calT^{(1)},Y^{(1)}),\ldots,(X^{(N)},\calT^{(N)},Y^{(N)})$, and a test example $(X,\calT,Y)$, such that all $(X^{(i)},\calT^{(i)},Y^{(i)})$ and $(X,\calT,Y)$ are exchangeable. Given a failure probability $\alpha\in(0,1)$, design conformal intervals $\calC_t(X)\subseteq\setR$ for the unknown observations $Y_t$, such that:
\begin{equation}\label{eq:coverage_guarantee}
\P\left(\forall t\in\calT: Y_t\in\calC_t(X)\right)\geq1-\alpha.    
\end{equation}
The intervals $\calC_t(X)$ may also depend on the dataset $D$ as well as the parameters $N$ and $\alpha$.
\end{problem}

% \begin{figure*}[ht]
% \vskip 0.1in
% \begin{center}
% \centerline{\includegraphics[width=\linewidth]{images/main_paper/internal_study_results.pdf}}
% \caption{We compare the mean coverage and mean interval width of our baseline and conformal prediction bands for hippocampal- and ventricular-volume trajectories. Error bars denote the 95th percentile of the metrics across 10 data splits. The baseline predictors (solid bars) either exceed the desired coverage by a noticeable margin (DKPG, DME) or fall short of it (DQR, DRMC, Bootstrap). In contrast, all conformalized predictors (striped bars) achieve the nominal coverage while also maintaining relatively tight intervals.}
% \label{fig:internal_study_results}
% \end{center}
% \vskip -0.1in
% \end{figure*}

\vspace{-0.3cm}
\subsection{Conformal Prediction Bands for Randomly-Timed Trajectories}\label{subsec:Conformal_Prediction Bands_for_Irregular_Trajectories}
\vspace{-0.3cm}

In this subsection, we introduce conformal prediction for randomly-timed trajectories, under the setting of Subsection~\ref{subsec:Problem_Formulation}. Inspired by the conformal methods in \cite{Yu2023,cleaveland2024time} for fixed-time trajectories, we design a prediction band that guarantees covering the unknown trajectory $Y$ with arbitrary confidence (see \eqref{eq:coverage_guarantee}). 

We focus on the standard tractable variant of conformal prediction, referred to as split conformal prediction \cite{Papadopoulos2008}. Specifically, we start by splitting the dataset $D$ into a training set $D_{\train}$ and a calibration set $D_{\calib}$. Let $\calI_{\train}:=\{i:(X^{(i)},\calT^{(i)},Y^{(i)})\in D_{\train}\}$ and $\calI_{\calib}:=\{i:(X^{(i)},\calT^{(i)},Y^{(i)})\in D_{\calib}\}$ denote the corresponding sets of indices. Employing the dataset $D_{\train}$, we train an arbitrary model that maps the input $X$ to predictions $\Yhat_t$ of the future observations $Y_t$ (see Appendix~\ref{sup:predictors} for examples of predictive models). Moreover, let $\Yhat_t^{(i)}$ denote the estimate of $Y_t^{(i)}$ returned by the learned predictor. The idea is to leverage the data in $D_{\calib}$ and the corresponding predictions $\Yhat_t^{(i)}$ to transform the estimates $\Yhat_t$ into intervals that ensure the guarantee \eqref{eq:coverage_guarantee}. To this end, we define the normalized \textit{nonconformity scores}: 
\begin{equation}\label{eq:nonconformity_scores}
    R^{(i)} = \max_{\substack{t\in\calT^{(i)}}}\left\{\frac{|Y_{t}^{(i)}-\Yhat_{t}^{(i)}|}{\sigma(\Yhat_t^{(i)})}\right\},
\end{equation}
where $\sigma(\cdot):\setR\to\setR_+$ is an arbitrary normalizing function, for all $i\in\calI_{\calib}$. A natural way of defining the function $\sigma(\cdot)$ is by leveraging some notion of predictive standard deviation (std). In cases of predictors with inherent predictive stds, the function $\sigma(\cdot)$ can be defined accordingly (see Appendix~\ref{sup:predictors} for examples). Otherwise, the factors $\sigma(\Yhat_t^{(i)})$ can be computed from the given data, employing ideas from \cite{Yu2023,cleaveland2024time}. In the following theorem, we use the scores in \eqref{eq:nonconformity_scores} to design a prediction band composed of intervals $\calC_t(X)\subseteq\setR$ that ensure the guarantee \eqref{eq:coverage_guarantee}. The proof of the theorem is given in Appendix~\ref{sec:Proof_of_Theorem_1}.

\begin{theorem}[Conformal Prediction Bands for Randomly-Timed Trajectories]\label{theorem}
Fix a failure probability $\alpha\in(0,1)$. Let $\Yhat_{t}$ be the prediction of the future observation $Y_t$ at time point $t$. Consider the nonconformity scores $R^{(i)}$ defined as in \eqref{eq:nonconformity_scores}, for any normalizing function $\sigma(\cdot)$. Then, if $R$ is the $\ceil{(|D_{\calib}|+1)(1-\alpha)}$\footnote{The notation $\ceil{\cdot}$ represents the ceiling function. \label{fn:ceiling}}-th smallest value of the set $\{R^{(i)}: i\in\calI_{\calib}\}\cup\{\infty\}$, the guarantee \eqref{eq:coverage_guarantee} holds with intervals $\calC_t(X):=[-R\sigma(\Yhat_t)+\Yhat_t,\Yhat_t+R\sigma(\Yhat_t)]$, $\forall t$.
\end{theorem}

Notice in Theorem~\ref{theorem} that the normalizing factors $\sigma(\Yhat_t)$ are leveraged to scale the score $R$ into $R\sigma(\Yhat_t)$ across different time points $t$. For the intervals $\calC_t(X)$
to be bounded, we need to have $\ceil{(|D_{\calib}|+1)(1-\alpha)}\leq|D_{\calib}|$. We note that, while the set of time points $\calT$ is a priori unknown, we can design a prediction band for $Y$ by deriving the intervals $\calC_t(X)$ for all time points $t\in\{1,\ldots,T\}$, for a user-defined maximum time $T\in\setN_+$, and interpret the intervals at time points outside $\calT$ as interpolations among the true time points (i.e., those contained in $\calT$). For example, in  Figure~\ref{fig:example_bands}, the conformal prediction band is guaranteed to cover the true observations (black bullets) at the corresponding (a priori unknown) time points.

% \begin{figure*}[ht]
% \vskip 0.1in
% \begin{center}
% \centerline{\includegraphics[width=0.9\linewidth]{images/main_paper/internal_study_results.pdf}}
% \caption{We compare the mean coverage and mean interval width of our baseline and conformal prediction bands for hippocampal- and ventricular-volume trajectories. Error bars denote the 95th percentile of the metrics across 10 data splits. The baseline predictors (solid bars) either exceed the desired coverage by a noticeable margin (DKPG, DME) or fall short of it (DQR, DRMC, Bootstrap). In contrast, all conformalized predictors (striped bars) achieve the nominal coverage while also maintaining relatively tight intervals.}
% \label{fig:internal_study_results}
% \end{center}
% \vskip -0.1in
% \end{figure*}

\begin{brem}\label{remark}
The nonconformity scores in \eqref{eq:nonconformity_scores} are inspired by the conformal methods in \citep{Yu2023,cleaveland2024time}, which guarantee coverage for fixed-time trajectories, where the sets of time points $\calT$ and $\calT^{(i)}$, $i=1,\ldots,N$, are identical and known a priori. However, in the context of biomarker trajectory prediction, observations are collected at random time points for each subject, as depicted in Figure~\ref{fig:irregular_patient_data}. Therefore, the application of the approach from \citep{Yu2023,cleaveland2024time} is hindered by the lack of prior knowledge of the time points  at which observations are collected for the test subject. More details on why these previous approaches cannot be used in our setting are given in~\ref{subsec:conformal_literature}. Our method removes this assumption and guarantees coverage of randomly-timed trajectories (see Theorem~\ref{theorem}).
\end{brem}

Beyond the coverage guarantee \eqref{eq:coverage_guarantee}, our conformal prediction method provides us with a distribution-agnostic framework for obtaining uncertainty-calibrated variants of arbitrary trajectory predictors. In the following section, we demonstrate the applicability of our approach to several standard and state-of-the-art predictive models.

\section{Prediction of Brain Biomarker Trajectories}
\label{sec:Prediction_Brain_Biomarker}
\vspace{-0.3cm}

\begin{figure*}[t]
\vskip 0.1in
\centering
% Bottom plot
\begin{subfigure}[t]{\linewidth}
    \centering
    \includegraphics[width=\linewidth]{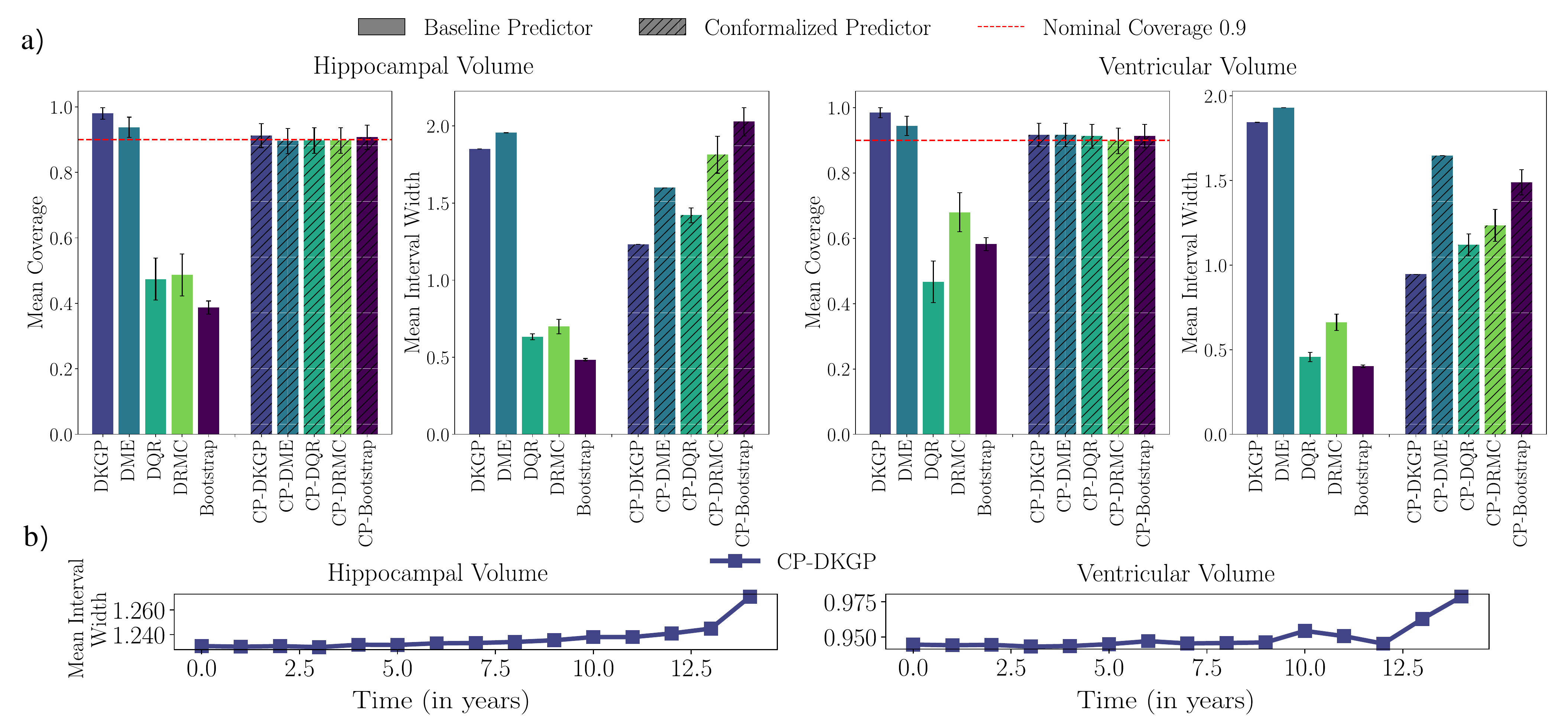}
\end{subfigure}
\caption{a) We compare the mean coverage and mean interval width of our baseline and conformal prediction bands for hippocampal- and ventricular-volume trajectories. Error bars denote the 95th percentile of the metrics across 10 data splits. The baseline predictors (solid bars) either exceed the desired coverage by a noticeable margin (DKPG, DME) or fall short of it (DQR, DRMC, Bootstrap). In contrast, all conformalized predictors (striped bars) achieve the nominal coverage while also maintaining relatively tight intervals.
b) We show the temporal evolution of the mean interval width of the  prediction bands resulting from the conformalized DKGP predictor with $\alpha=0.1$ for hippocampal- and ventricular-volume trajectories. Notice that the bands become wider on average over time, reflecting the expected growth in uncertainty as the prediction horizon extends. Reported interval widths are on the standardized scale.}
\label{fig:merged_interval_results}
\end{figure*}

\begin{figure}
    \vspace{-0.1in}
    \centering
    \includegraphics[width=0.5\linewidth] {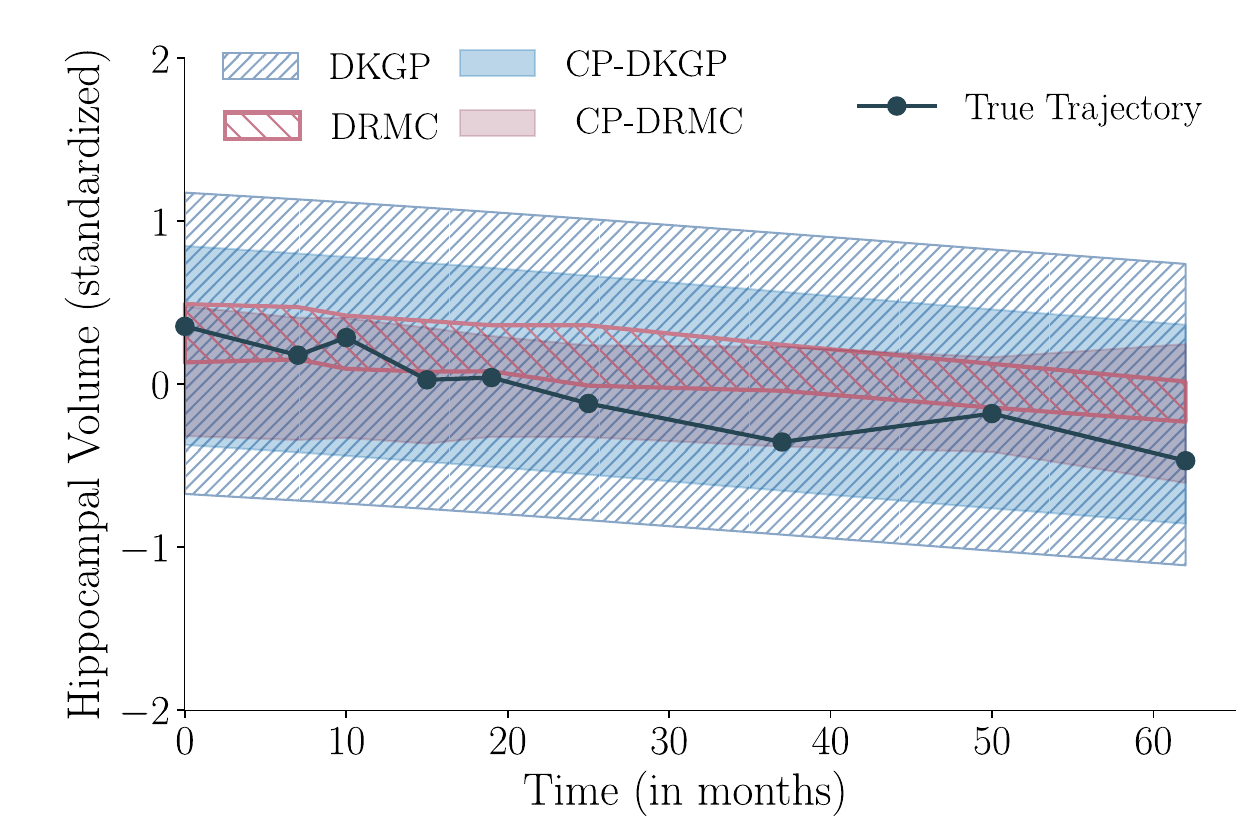}
    \caption{We compare baseline and conformal prediction bands obtained from the DKGP and DRMC predictors for a test subject's hippocampal-volume trajectories. Our conformalized DKGP predictor produces tighter bands than its baseline counterpart. Our conformalized DRMC predictor adjusts the respective baseline prediction band in order to achieve covering the true biomarker values at all time points.}
    \label{fig:example_bands}
    \vspace{-0.1in}
\end{figure}

In this section, we apply our conformal prediction method in a case study involving two well-established brain biomarkers. Particularly, we focus on predicting hippocampal- and ventricular-volume trajectories, both of which are critical for diagnosing and monitoring the progression of neurodegenerative diseases, such as Alzheimer's disease (AD) \cite{Apostolova2012}. For each biomarker, we use a dataset of $2,200$ samples with $(X^{(i)},\calT^{(i)},Y^{(i)})$ from subject $i$: i) $X^{(i)}$ are input features consisting of demographic and clinical covariates, such as sex, race, and clinical diagnosis, as well as volume measures from 145 brain regions, collected on the subject's first visit;
ii)  $\calT^{(i)}$ includes the set of time points of the subject's subsequent visits, relative to their first visit;
iii)  $Y^{(i)}$ is the output trajectory involving the corresponding biomarker values at the observed time points.
For our experiment, we combine the preprocessed and harmonized neuroimaging measures from two well-known longitudinal studies--- Alzheimer's Disease Neuroimaging Initiative (ADNI) \citet{ADNI} and Baltimore Longitudinal Study of Aging (BLSA) \citep{Ferrucci2008}---which focus on AD and Brain Aging, respectively. Details on the studies and our preprocessing pipelines can be found in Appendix~\ref{sup:datasets}.

To demonstrate the model-agnostic nature of our framework, we apply it to various standard and state-of-the-art predictors: i) Deep Kernel Gaussian Process (DKGP) model \cite{pmlr-v193-tassopoulou22a}, ii) Deep Mixed Effects (DME) model \cite{chung2019deepmixedeffectmodel}, iii) Deep Quantile Regression (DQR) model \citep{10.1257/jep.15.4.143}, iv) Deep Regression with Monte Carlo (DRMC) dropout model \cite{gal2016dropoutbayesianapproximationrepresenting}, and v) Bootstrap Deep Regression model \cite{Breiman1996}. Details on the architecture and training of each model can be referred to Appendix~\ref{subsec:Architectural_Design_and_Training}. We note that existing conformal prediction methods are designed for fixed-time settings and cannot be applied to randomly-timed trajectories, which precludes direct comparisons. We elaborate further in Appendix~\ref{subsec:conformal_literature}.

For each predictor, we present a comparative case study between its \textit{baseline} variant, which is trained on the entire given dataset, and its \textit{conformalized} variant, which is trained on a subset of the given dataset, as described in Subsection~\ref{subsec:Conformal_Prediction Bands_for_Irregular_Trajectories}. To design prediction bands for the baseline variants, we employ the inherent uncertainty measure provided by their predictive standard deviation (see Appendix~\ref{subsec:Predictive_Standard_Deviation} for details). By leveraging the same uncertainty measure, we define the normalizing function $\sigma(\cdot)$ employed in the design of our conformal prediction bands (see Theorem~\ref{theorem}). We note that while baseline bands provide a notion of uncertainty-calibrated prediction, the desired coverage might not be achieved when model assumptions are violated. 

In all experiments, we evaluate the performance of the baseline and conformal prediction bands in terms of mean coverage (i.e., the proportion of test trajectories entirely contained within the prediction bands) and mean interval width (across all test trajectories and corresponding time points). Specifically, we compute these metrics by averaging over $10$ random splits of the dataset into given data and test data, with the test set comprising 10\% of the total dataset. For the design of our conformal prediction bands, we perform an additional splitting of the given data into training and calibration sets. In all cases, the calibration set size is selected as 20\% of the given dataset (see Appendix~\ref{subsec:Calibration Set Size Selection} for details about this choice). Moreover, in this section, we consider a confidence level of 0.9 (i.e., $\alpha=0.1$), while additional confidence levels are explored in Appendix~\ref{subsec:Comparison_with_Baselines_for_Other_Confidence Levels}. 

In Figure~\ref{fig:merged_interval_results}a, we illustrate the mean coverage and mean interval width attained by all the baseline and conformalized predictors for the two biomarkers of interest. We observe that in the case of the DKGP and DME predictors, both the baseline and conformalized variants achieve the desired coverage. However, the mean coverage of the conformalized predictors is closer to 0.9, and the corresponding prediction bands are tighter, as indicated by their reduced mean interval width compared to the baselines. We also see that unlike the corresponding baseline variants, the conformalized DQR, DRMC, and Bootstrap predictors are able to achieve the desired coverage, confirming our coverage guarantee in \eqref{eq:coverage_guarantee}. Figure~\ref{fig:merged_interval_results}b shows the temporal evolution of the mean interval width for the conformalized DKGP predictor. We observe the intervals become gradually wider over time, which is expected since uncertainty typically increases as the prediction horizon increases. Similar behavior is observed for the remaining predictors in Appendix~\ref{subsec:Mean_Interval_Widths_over_Time}. In Figure~\ref{fig:example_bands}, we present baseline and conformal prediction bands obtained from the DKGP and DRMC predictors for a test subject's hippocampal-volume trajectories. 

\vspace{-0.3cm}
\section{Group-Conditional Application across Covariate Subpopulations}
\label{sec:Stratified_Evaluation}
\vspace{-0.3cm}
So far, we have developed and tested conformalized predictors of biomarker trajectories that guarantee coverage across the \textit{overall patient population}. However, these guarantees may not hold uniformly when applied to \textit{subpopulations} defined by demographic or clinical covariates. In particular, high-risk groups (e.g., individuals with MCI or genetic risk factors) and underrepresented demographic subgroups may be inadequately captured by population-based conformal bands, leading to miscalibrated uncertainty estimates. To address this, we employ a group-conditional variant of our approach, directly derived from the conformal framework in \cite{Vovk2005,Vovk2012}. We describe this method in Subsection~\ref{subsec:Group-Conditional Conformal Prediction for randomly-timed Trajectories} and evaluate it in our case study on hippocampal- and ventricular-volume trajectories in Subsection~\ref{subsec:Application to our Case Study}.

\vspace{-0.3cm}
\subsection{Group-Conditional Conformal Prediction for Randomly-Timed Trajectories}\label{subsec:Group-Conditional Conformal Prediction for randomly-timed Trajectories}
\vspace{-0.3cm}

In this subsection, we present a group-conditional variant of the conformal prediction method described in Subsection~\ref{subsec:Conformal_Prediction Bands_for_Irregular_Trajectories}. Specifically, we provide group-conditional conformal prediction bands for randomly-timed trajectories, following the Mondrian conformal prediction framework from \cite{Vovk2005,Vovk2012}.

Mondrian conformal prediction is a general approach for deriving group-conditional variants of conformalized predictors. To apply it to our setting of randomly-timed trajectories for distinct covariate subpopulations, consider a grouping function $G:\setR^d\to\calG$, where $\calG$ is a predefined set of covariate groups. The idea is to stratify the calibration data by group, and apply the conformal method from Subsection~\ref{subsec:Conformal_Prediction Bands_for_Irregular_Trajectories}
to each of the groups. In this way, we can derive group-conditional conformal bands that guarantee coverage within each covariate subpopulation. We formalize our result in the following corollary, adopting the notation from Section~\ref{sec:Conformal_Prediction_for_Irregular_Time_Series}. The proof is omitted, as it directly follows from the application of Theorem~\ref{theorem} across all subpopulations, similar to the proof of \citep[Proposition 3]{Vovk2012}.

\begin{corollary}[Group-Conditional Conformal Prediction Bands for Randomly-Timed Trajectories]\label{lemma}
Fix a failure probability $\alpha\in(0,1)$. Let $\Yhat_{t}$ be the prediction of the future observation $Y_t$ at time point $t$. Consider the nonconformity scores $R^{(i)}$ defined as in \eqref{eq:nonconformity_scores}, for any normalizing function $\sigma(\cdot)$. For each $g\in\calG$, let $D_{\calib,g}:=\{(X^{(i)},\calT^{(i)},Y^{(i)})\in D_{\calib}: G(X^{(i)})=g\}$ be the corresponding subset of $D_{\calib}$ and $\calI_{\calib,g}:=\{i\in\calI_{\calib}: G(X^{(i)})=g\}$ the corresponding set of indices. Moreover, for each $g\in\calG$, let $R_g$ denote the $\ceil{(|D_{\calib,g}|+1)(1-\alpha)}$\footref{fn:ceiling}-th smallest value of the set $\{R^{(i)}: i\in\calI_{\calib,g}\}\cup\{\infty\}$. Then, for every $g\in\calG$, we have:
\begin{equation*}\label{eq:group_conditional_guarantee}
    \P\left(\forall t\in\calT: Y_t\in\calC_t(X)\,|\,G(X)=g\right)\geq1-\alpha,
\end{equation*}
with $C_t(X)=[-R_{G(X)}\sigma(\Yhat_t)+\Yhat_t,\Yhat_t+R_{G(X)}\sigma(\Yhat_t)]$, $\forall t$.
\end{corollary}
Simply put, the above corollary guarantees that, given a test example from a known covariate group $g$, we can derive conformal prediction intervals with guaranteed coverage of $1-\alpha$ for the unknown trajectory $Y$, for any $\alpha\in(0,1)$.

\vspace{-0.3cm}
\subsection{Application across Heterogeneous Demographic and Clinical Covariate Groups}\label{subsec:Application to our Case Study}
\vspace{-0.3cm}

In this subsection, we test our group-conditional conformal prediction method in the case study of hippocampal- and ventricular-volume trajectories, described in Section~\ref{sec:Prediction_Brain_Biomarker}. Specifically, we design group-conditional conformal prediction bands for a total of five population stratifications, each based on a distinct demographic or clinical covariate. Among demographic factors, we consider sex, race, and education level, with the following respective subpopulations: i) females and males, ii) Asians, Blacks, and Whites, and iii) subjects with less than 16 years of education ($\text{Edu}<16$) and more than 16 years of education ($\text{Edu}>16$). As for clinical covariates, we consider diagnosis and APOE4 allele status, a genetic risk factor associated with Alzheimer's disease. As noted in Section~\ref{sec:Prediction_Brain_Biomarker}, our diagnostic composition divides subjects into: Cognitively Normal (CN), Mildly Cognitively Impaired (MCI), and subjects diagnosed with Alzheimer's disease (AD). Moreover, subjects are classified based on their APOE4 status as non-carriers, heterozygotes and homozygotes depending on whether they have zero, one, or two copies of the APOE4 allele, respectively.

\begin{figure}
    \vspace{-0.1in}
    \centering
    \includegraphics[width=\linewidth]{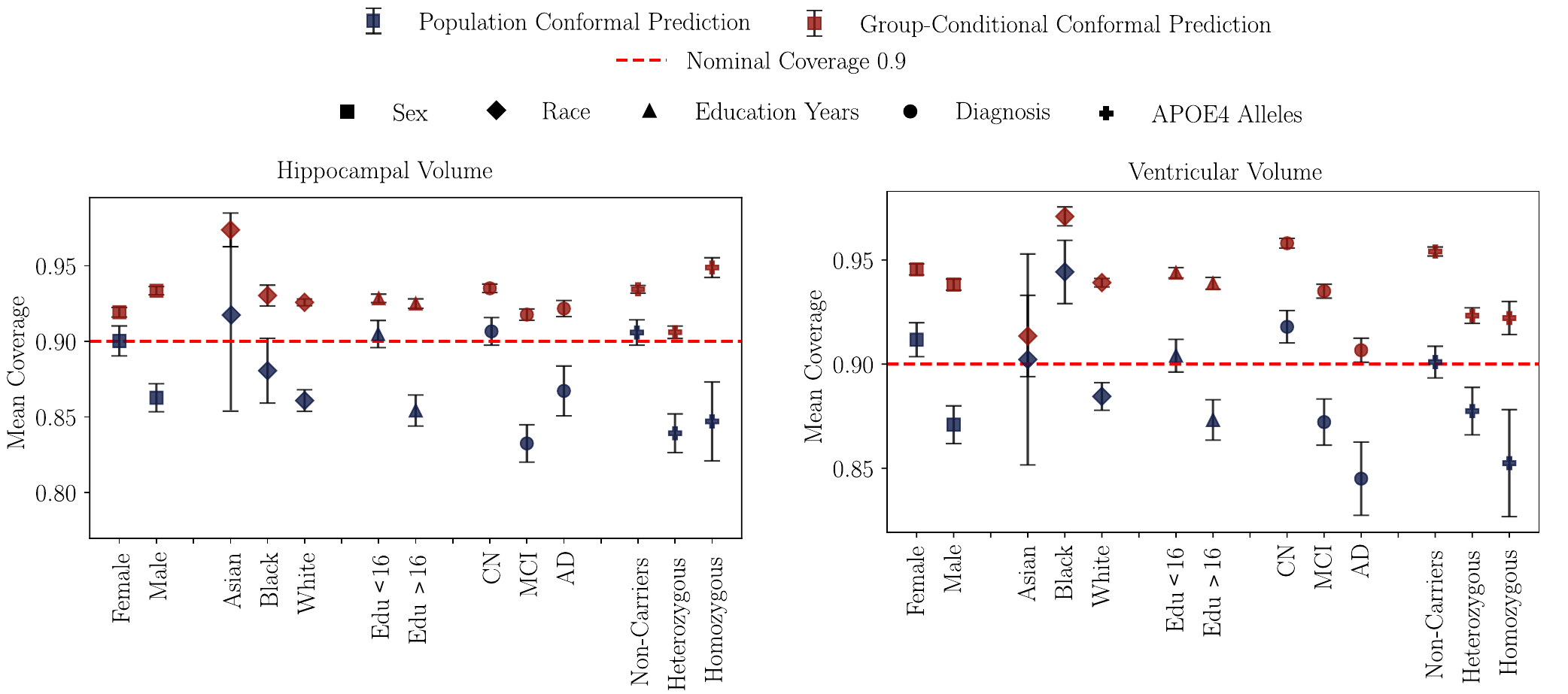}
   \caption{We compare the mean coverage of \textit{population} and \textit{group-conditional} conformal prediction bands for hippocampal- and ventricular-volume trajectories. Error bars denote the 95th percentile of the metrics across 10 data splits. Results are presented across five population stratifications based on individual covariates. For most covariate groups, the population conformalized predictors fail to achieve the desired coverage, while the group-conditional conformalized predictors consistently attain the desired confidence level.}
    \label{fig:stratified}
    \vspace{-0.1in}
\end{figure}

To demonstrate the impact of our group-conditional conformal method, we compare it with the approach from Subsection~\ref{subsec:Conformal_Prediction Bands_for_Irregular_Trajectories} for the five population stratifications. Herein, we refer to the latter method as \textit{population conformal prediction}, owing to its coverage guarantees across the overall population. Our experiments focus on the setup of Section~\ref{sec:Prediction_Brain_Biomarker}, employing the state-of-the-art Deep Kernel Gaussian Process model from \citep{pmlr-v193-tassopoulou22a}, as well as the combined cohort of the ADNI and BLSA studies. For the composition of the dataset across all five population strata, we refer the readers to Table~\ref{table:demographics_and_clinical} in Appendix~\ref{sup:datasets}. We conduct experiments for  10 distinct splits of the data into training, calibration, and test sets, as detailed in Section~\ref{sec:Prediction_Brain_Biomarker}. For each of the splits, we first learn a model from the corresponding training data, and then design: i) a single population conformalized predictor, and ii) five group-conditional conformalized predictors for the respective population stratifications. For the group-conditional conformal prediction bands, the test data are stratified by group. The mean coverage and mean interval width are computed by averaging over all 10 data splits. Results are presented  for a confidence level of 0.9, while additional values are  in Appendix~\ref{sup:extended_stratif}.  

% Figure~\ref{fig:stratified} depicts the mean coverage of the population and group-conditional conformal prediction bands for hippocampal-volume trajectory. We observe that the population conformalized predictors (in blue) fail to achieve the nominal coverage across most covariate groups. Significantly deviation and undercoverage is noted for minority populations, such as Asians, subjects diagnosed with AD, and APOE4 homozygotes. On the other hand, the group-conditional conformalized predictors (in red) achieve the nominal coverage for all covariate groups. Additional figures with the mean interval width across all five population stratifications are provided in Appendix~~\ref{sup:extended_stratif}. 

Figure~\ref{fig:stratified} shows the mean coverage achieved by population- and group-conditional conformal predictors across five covariate stratifications for hippocampal-volume trajectories.  We observe that the population conformalized predictors (in blue) fail to achieve the nominal coverage across most covariate groups in several cases—including clinically high-risk or underrepresented groups such as individuals diagnosed with Mild Cognitive Impairment (MCI), APOE4 homozygotes, and Asian subjects. On the other hand, while group-conditional predictors (red) consistently attain the nominal 90\% coverage across all subpopulations. 

Our empirical results confirm that our group-conditional conformal prediction bands guarantee reliable coverage of biomarker trajectories within diverse covariate subpopulations. This underscores the potential of our approach for clinical deployment, especially in guiding early intervention for high-risk groups, such as MCI or APOE4 homozygotes. In the next section, we illustrate how our conformal bands can be directly leveraged in a downstream clinical task focused on identifying subjects at high risk of progression to AD.

\vspace{-0.3cm}
\section{Clinical Application: Identification of High-Risk Alzheimer's Progressors}
\label{sec:clinical_applications}
\vspace{-0.3cm}

In this section, we demonstrate the clinical utility of the proposed uncertainty-calibrated prediction method for identifying individuals at high risk of progression to Alzheimer's disease. Specifically, we employ our conformal bands to stratify MCI subjects into MCI-stable (who will not progress to AD) and MCI-progressors (who will progress to AD). In clinical practice, the rate of change (RoC) of certain biomarkers between the subject's first clinical visit and a subsequent clinical visit has been proven significant for early detection of AD and cohort enrichment in relevant clinical trials \cite{Wang2020,Rajan2023,Maheux2023Forecasting,Schaap2024}. In our setting of biomarker trajectory prediction, we define the \emph{predicted rate of change} ($\operatorname{\hat{RoC}}$) of subject $i$ between an initial time point $t_0$ and a future time point $t_N$ as:
\begin{equation}
\operatorname{\hat{RoC}}^{(i)}
\;=\;
\frac{\hat{Y}^{(i)}_{t_N} - Y^{(i)}_{t_0}}{t_{N} - t_{0}}.
\label{eq:roc}
\end{equation}

While the predicted rate of change is widely used as a progression marker, relying on trajectory predictions alone can underestimate risk in uncertain cases. We therefore introduce an uncertainty-calibrated risk score that quantifies the worst-case rate of change with high probability, which we call \emph{rate-of-change bound} ($\operatorname{RoCB}$). Let $L^{(i)}_t$ and $U^{(i)}_t$ denote the lower and upper bound, respectively, of a prediction interval obtained for subject $i$ at time point $t$ with a given method (e.g., a baseline method or our conformal prediction method). To account for the different behaviors of distinct biomarkers, the $\operatorname{RoCB}$ of subject $i$ between time points $t_0$ and $t_N$ is defined as:
\begin{equation}
\operatorname{RoCB}^{(i)}
\;=\;
\begin{cases}
\dfrac{L^{(i)}_{t_{N}}-Y^{(i)}_{t_{0}}}{t_{N} - t_{0}}, & \text{if biomarker decreases with progression}, \\[1.2em]
\dfrac{U^{(i)}_{t_{N}} - Y^{(i)}
_{t_{0}}}{t_{N} - t_{0}}, & \text{if biomarker increases with progression}.
\end{cases}
\label{eq:broc}
\end{equation}

Note that the $\operatorname{RoCB}$ provides a notion of an uncertainty-calibrated risk score. In particular, for decreasing biomarkers, such as hippocampal volume, the worst-case decline is captured via the lower bound of the prediction interval at time point $t_N$, whereas for increasing biomarkers, such as ventricular volume, the upper bound of the interval is used.

\begin{table}
  \centering
  %\scriptsize
  \setlength{\tabcolsep}{5pt}
  \caption{Youden-optimal discrimination of MCI-stable subjects and MCI progressors to AD using the $z$-standardized predicted rate of change ($\operatorname{\hat{RoC}}$) and the rate-of-change bound ($\operatorname{RoCB}$) with 95\% bootstrap confidence intervals. 
  Since hippocampal volume is a decreasing biomarker, $\operatorname{RoCB}$ here corresponds to the lower rate-of-change bound.}
  \label{tab:mci_wcrc_lower}
  \begin{tabular}{llcccc}
    \toprule
    \textbf{Method} & \textbf{Metric} & $\boldsymbol{\tau^{\star}}$
    & \textbf{Precision} & \textbf{Recall} & \textbf{F$_1$} \\
    \midrule
    DRMC     & $\operatorname{\hat{RoC}}$  & $-0.006$ & \textbf{0.436} $\pm$ 0.022 & 0.671 $\pm$ 0.058 & 0.528 $\pm$ 0.023 \\
             & $\operatorname{RoCB}$ & $-0.012$ & 0.403 $\pm$ 0.022 & \textbf{0.884} $\pm$ 0.058 & \textbf{0.553} $\pm$ 0.023 \\
    CP--DRMC & $\operatorname{\hat{RoC}}$  & $-0.006$ & \textbf{0.432} $\pm$ 0.022 & 0.740 $\pm$ 0.095 & 0.546 $\pm$ 0.024 \\
             & $\operatorname{RoCB}$ & $-0.020$ & 0.395 $\pm$ 0.022 & \textbf{0.915} $\pm$ 0.095 & \textbf{0.552} $\pm$ 0.024 \\
    \bottomrule
  \end{tabular}
\end{table}

In the following experiments, we focus on hippocampal volume which is a key neurodegenerative biomarker related to Alzheimer's disease progression. We evaluate each subject's predicted rate of change as a baseline risk score and compare it with the proposed rate-of-change bound. Using each of these risk scores, we estimate a threshold $\tau^{\star}$ that maximizes discrimination between MCI-stable subjects and MCI-progressors using Youden's index \cite{ruopp2008youden}. Subjects with values below or equal to this threshold are classified as high-risk, whereas those that exceed it are classified as stable. We note that the cohort includes 462 MCI-stable participants (mean follow-up 22.5 months; range 0--153 months) and 258 MCI-progressors (mean 23.5 months; range 0--142 months).

\begin{figure*}[t]
\vskip 0.1in
\centering
% Bottom plot
\begin{subfigure}[t]{\linewidth}
    \centering
    \includegraphics[width=\linewidth]{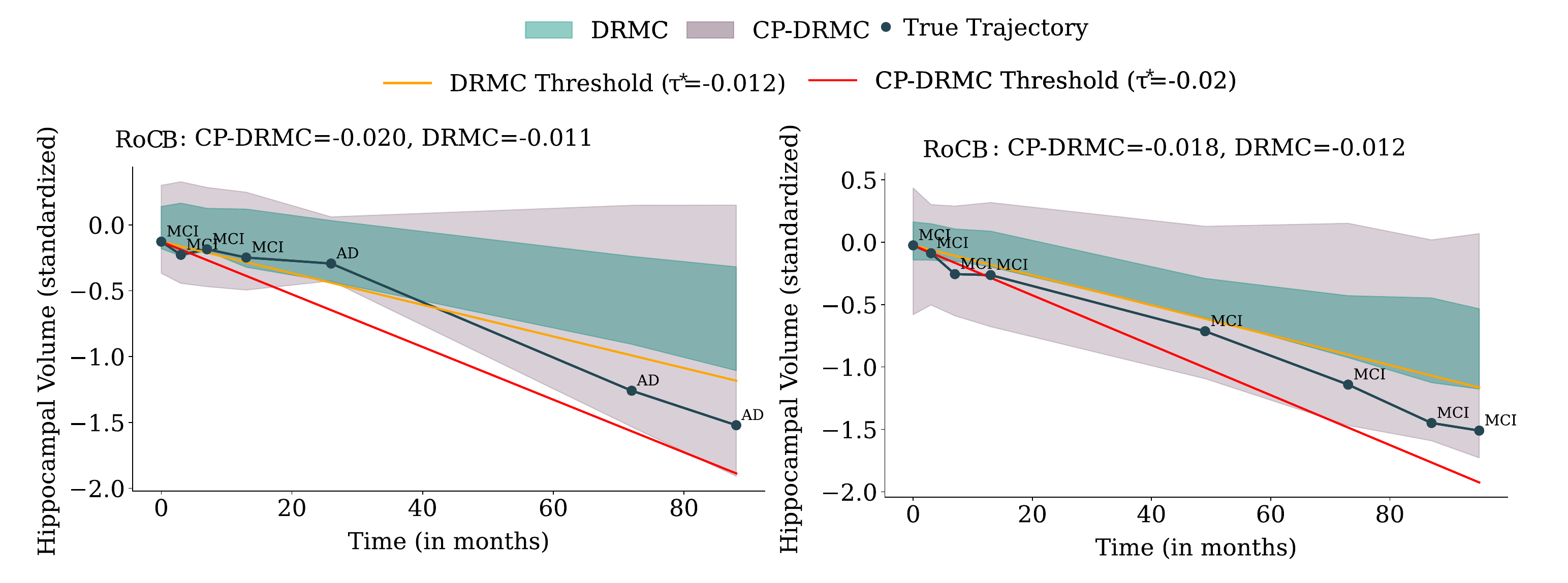}
\end{subfigure}
\caption{Hippocampal-volume trajectories (black) and prediction bands for two MCI participants. Left: Progressor from MCI to AD. The DRMC band (green) fails to cover the true trajectory and yields a rate-of-change bound of $-0.011$ (SD/month)—above the Youden-optimal threshold—leading to the subject being misclassified as MCI-stable. Conformal calibration (CP-DRMC, lilac) widens the prediction band and produces $\operatorname{RoCB}=-0.020$ (SD/ month)—above the Youden-optimal threshold—classifying the subject correctly as an MCI-progressor. Right: MCI-stable subject. DRMC erroneously predicts progression to AD, whereas the wider CP-DRMC band raises the $\operatorname{RoCB}$ above the respective threshold, preventing a false detection. Together these examples suggest that conformalization can rescue under-scored progressors while also reducing false positives among stable cases. SD stands for standard deviation.}
\vskip -0.1in
\label{fig:example_clinical}
\end{figure*}

To evaluate the clinical value of our conformal bands, we compare the discriminative performance of DRMC (with its baseline uncertainty measures---see Section~\ref{sec:Prediction_Brain_Biomarker}) and its conformalized counterpart (CP--DRMC) using the Youden-optimal threshold computed with both the predicted rate of change and the rate-of-change bound. As shown in Table~\ref{tab:mci_wcrc_lower}, our uncertainty-calibrated risk score substantially improves recall for both predictors. In particular, we observe an increase in recall from 67.1\% ($\operatorname{\hat{RoC}}$) to 88.4\% ($\operatorname{RoCB}$) for the DRMC predictor, and from 74.0\% to 91.5\% for its conformalized version. These increases demonstrate the benefit of incorporating predictive uncertainty into the standard risk score. Notice also that the recall of the conformalized predictors' rate-of-change bound is higher than that of the baseline predictors. This observation suggests that conformal calibration not only produces well-calibrated prediction bands but also improves risk evaluation. Although precision decreases when using $\operatorname{RoCB}$ instead of $\operatorname{RoC}$, the F$_1$ score remains similar, implying that the increased sensitivity does not come at the cost of overall discrimination. Table \ref{tab:supp_mci_wcrc_full_ci} in Appendix~\ref{sup:clinical_applications} presents similar results for the remaining predictors of Section~\ref{sec:Prediction_Brain_Biomarker}.

To further illustrate how conformal prediction improves clinical decision making, Figure~\ref{fig:example_clinical} presents two representative hippocampal-volume trajectories comparing the DRMC predictor and its conformalized counterpart. On the left, an MCI-progressor is misclassified by DRMC; particularly, the model’s overconfident prediction band fails to cover the true biomarker trajectory, and the resulting $\operatorname{RoCB}$ lies above the Youden-optimal threshold, leading to a missed detection. In contrast, our conformalized DRMC predictor produces a wider  prediction band that lowers the $\operatorname{RoCB}$ sufficiently for the subject to be correctly classified as an MCI-progressor. On the right, we show a stable MCI subject incorrectly identified as high-risk by DRMC due to an overly steep $\operatorname{\hat{RoC}}$. The conformalized predictor raises $\operatorname{RoCB}$ to a larger value and avoids false identification.

\vspace{-0.3cm}
\section{Discussion on Clinical Impact}
\vspace{-0.3cm}
Our framework brings distribution-free, uncertainty-calibrated predicting to biomarker trajectories observed at \emph{randomly-timed} visits---a pervasive reality in longitudinal clinical studies. By conformalizing trajectory predictions, we equip downstream decisions (e.g., early intervention, enrichment for MCI or APOE4 subgroups) with formal coverage guarantees at subject-specific time points. Importantly, our approach is model-agnostic and can conformalize any biomarker predictor to enhance reliability in precision trial design \cite{Cummings2019}, where slope-based metrics guide inclusion to the clinical trial \cite{Maheux2023Forecasting}. Our rate-of-change bound ($\operatorname{RoCB}$) uses the conformal band to favor conservative, safety-oriented screening without parametric uncertainty assumptions. Both population- and group-conditional calibration improve trustworthiness in high-risk or underrepresented subgroups.

\vspace{-0.3cm}
\section{Limitations and Future Work}
\vspace{-0.3cm}
Our paper points to several future research directions. First, we could consider deriving prediction bands for trajectories of multiple biomarkers simultaneously as an extension of our conformal method to multivariate predictors. It would also be interesting to explore adaptive variants of our framework for predictors that provide updated subject-specific biomarker estimates over time.  Additionally, our current coverage guarantees apply when the test and calibration data are drawn from the same underlying distribution (i.e., they are exchangeable). When this assumption is violated---such as when applying the method to new cohorts from different sites, imaging protocols, or demographic compositions---the coverage guarantees may no longer hold. Addressing such distribution shifts remains an open problem.  Furthermore, it would be interesting to expand our group-conditional conformalized predictors to account for groups defined by combinations of multiple covariates (e.g., Black females aged 55-65), which could have a substantial impact on real-world healthcare applications.

\section*{Acknowledgement}
Vasiliki Tassopoulou acknowledges support from the NIH U24NS130411 RF1AG054409 grant, the NIA contract ZIA-AG000191. Charis Stamouli and George J. Pappas acknowledge support from ASSET (AI-Enabled Systems: Safe, Explainable and Trustworthy) Center.

\bibliographystyle{unsrtnat}
\bibliography{manuscript}

\newpage
\appendix
\onecolumn
\section*{Technical Appendices}

\section{Exchangeability}\label{sec:Exchangeability}
Exchangeability of a sequence of random variables implies that their joint probability distribution remains unchanged under any permutation of the variables. This condition is milder than the standard one of independent and identically distributed random variables. Below we provide the formal definition of exchangeability from \cite{Shafer2008}.

\begin{definition}
The variables $Z^{(1)},\ldots,Z^{(N+1)}$ are exchangeable if for every permutation $\pi(\cdot)$ of the integers $1,\ldots,N+1$, the variables $W^{(1)},\ldots,W^{(N+1)}$,
where $W^{(i)}=Z^{(\pi(i))}$, have the same joint probability distribution as $Z^{(1)},\ldots,Z^{(N+1)}$.
\end{definition}

\section{Proof of Theorem~\ref{theorem}}\label{sec:Proof_of_Theorem_1}

For simplicity of presentation, we introduce the notation $(X^{(N+1)},\calT^{(N+1)},Y^{(N+1)})$ to represent the test sample $(X,\calT,Y)$ and the notation $R^{(N+1)}$ to represent the random variable $R$ defined in the theorem statement. Since the random variables $(X^{(i)},\calT^{(i)},Y^{(i)})$, $i\in\{1,\ldots,N+1\}$, are exchangeable by assumption, the random variables $(X^{(i)},\calT^{(i)},Y^{(i)})$, $i\in\calI_{\calib}\cup\{N+1\}$, are also exchangeable. Somewhat abusing notation, we define the nonconformity score function:
\[
    R(X,\calT,Y) = \max_{t\in\calT}\left\{\frac{|Y_t-\Yhat_t|}{\sigma(\Yhat_t)}\right\},
\]
that applied to each datapoint $(X^{(i)},\calT^{(i)},Y^{(i)})$, $i\in\calI_{\calib}\cup\{N+1\}$, yields the corresponding score $R^{(i)}$. Conditioned on the dataset $D_{\train}$, the function $R(\cdot)$ is deterministic. Therefore, we conclude that conditioned on $D_{\train}$, the random variables $R^{(i)}$, $i\in\calI_{\calib}\cup\{N+1\}$, are exchangeable. Hence, from \citep[Lemma 1]{Tibshirani2019} we obtain the conditional property:
\begin{equation}\label{theorem1_1}
    \Prob\left(\max_{t\in\calT}\left\{\frac{|Y_t-\Yhat_t|}{\sigma(\Yhat_t)}\right\}\leq R\,\bigg|\,D_{\train}\right)\geq1-\alpha,
\end{equation}
where $R$ is defined as in the theorem statement. By marginalizing over $D_{\train}$, \eqref{theorem1_1} yields the unconditional property:
\begin{equation}\label{theorem1_2}
    \Prob\left(\max_{t\in\calT}\left\{\frac{|Y_t-\Yhat_t|}{\sigma(\Yhat_t)}\right\}\leq R\right)\geq1-\alpha.
\end{equation}
The guarantee \eqref{eq:coverage_guarantee} directly follows from rewriting \eqref{theorem1_2} as:
\begin{equation*}
    \Prob\left(\forall t\in\calT: |Y_t-\Yhat_t|\leq R \sigma(\Yhat_t)\right)\geq1-\alpha
\end{equation*}
and setting $\calC_t(X)=[-R\sigma(\Yhat_t)+\Yhat_t,\Yhat_t+R\sigma(\Yhat_t)]$, for all $t$.

$\qed$

\section{Discussion on Clinical Impact}
Our conformal prediction framework provides prediction bands that enable clinicians to perform informed prognosis and decision-making with greater reliability. This is important as predictive models are increasingly applied in healthcare for both patient management and drug development. In the latter case, \citet{Cummings2019} highlighted the need for AI-informed clinical trials,  referred to as precision trial design. Along these lines, \citet{Maheux2023Forecasting} evaluates a predictive model for biomarker trajectories, in the context of AD, where derived measures—such as biomarker rate of change—serve as quantitative indicators of whether a subject is likely to progress to the disease during the clinical trial. This assessment ultimately informs decisions on subject inclusion in the drug administration process. By conformalizing such predictors, either population level or calibrated in covariate groups, such as MCI subjects, we provide confidence bands alongside predictions, thus increasing the reliability of these critical clinical decisions.

\section{Clinical Datasets and Preprocessing}
\label{sup:datasets}
Our data consists of neuroimaging and demographic measures taken from subjects in the iSTAGING consortium \citep{habes2021brain}. Specifically, the neuroimaging measures are the 145 anatomical brain Regions of Interest (ROI) volumes (119 ROIs in gray matter, 20 ROIs in white matter and 6 ROIs in ventricles)  extracted using a multi‐atlas label fusion method \citep{Doshi2016}. Phase-level harmonization is applied on these 145 ROI volumes to remove site effects \citep{pomponio2020harmonization}. 
To train, calibrate and test the conformalized predictors we use data from two cohorts: the Alzheimer\'s Disease Neuroimaging Initiative \citep{ADNI}, which is a public-private collaborative longitudinal cohort study and has recruited participants categorized as Cognitively Normal (CN), Mildly Cognitively Impaired (MCI) and diagnosed with Alzheimer's Disease (AD) through 4 phases (ADNI1, ADNIGO and ADNI2) \citep{Weiner2017}. We also use Baltimore Longitudinal Study of Aging (BLSA) follows participants who are cognitively normal at enrollment with imaging and cognitive exams since 1993 \citep{Ferrucci2008}.

We also extract from iSTAGING cohort: the OASIS \cite{10.1162/jocn.2009.21407}, The Wisconsin Registry for Alzheimer’s Prevention (WRAP) study \cite{johnson2018wisconsin}, the Australian Imaging, Biomarker, and Lifestyle (AIBL) study \citep{lamontagne2019oasis}, the Coronary Artery Risk Development in Young Adults (CARDIA) \citep{FRIEDMAN19881105}, the PreventAD \citep{Tremblay-Mercier2021} and PENN. 

For the clinical variables, we utilize Age at Baseline, Sex, Years of Education, and APOE4 Allele status, the latter being a known risk factor for Alzheimer's Disease. Diagnostic categories were designated as Cognitively Normal (CN), Mild Cognitive Impairment (MCI), and Alzheimer's Disease (AD). Subjects diagnosed with alternative forms of dementia, such as Lewy Body Dementia and Frontotemporal Dementia, were excluded from the study. These exclusions are minimal (less than 10 subjects) and did not impact the overall sample size. After filtering, our dataset consists of 2200 subjects. Furthermore, Years of Education was dichotomized: subjects with more than 16 years of education were coded as '1', while those with 16 years or fewer were coded as '0'. Also, duplicate acquisitions within the same month are discarded.

\begin{table}[t]
  \caption{Demographic and clinical characteristics of the clinical studies. The joint cohort of ADNI and BLSA comprises our base population, whereas the remaining cohorts represent external populations. For the total time of observation and the age, we report the mean and standard deviation over all subjects contained in each study. CN: Cognitively Normal, MCI: Mild Cognitive Impairment, AD: Alzheimer's Disease.}
  \label{table:longitudinal_studies_detailed}
  \centering
  \begin{tabular}{lcccccccc}
    \toprule
    \textbf{Study} & \textbf{Subjects} & \textbf{Obs. Time (mo)} & \textbf{Males (\%)} & \textbf{Age} & \multicolumn{3}{c}{\textbf{Diagnosis (\%)}} \\
    \cmidrule(lr){6-8}
     &  &  &  &  & \textbf{CN} & \textbf{MCI} & \textbf{AD} \\
    \midrule
    ADNI+BLSA     & 2200 & 65 ± 39       & 53.2  & 75.4 ± 8.6  & 49.3  & 34.8  & 15.9 \\
    \bottomrule
  \end{tabular}
\end{table}

\begin{table}[t]
  \caption{Composition of the joint cohort of the ADNI and BLSA studies across five demographic and clinical covariates. }
  \label{table:demographics_and_clinical}
  \centering
  \begin{tabular}{lccc}
    \toprule
    \textbf{Covariate} & \textbf{Groups} & \textbf{Percentage of Subjects} & \textbf{Number of Subjects} \\
    \midrule
    \textbf{Sex} & Male & 52.89 & 1164 \\
    \textbf{Race} & White & 87.46 & 1924 \\
     & Black & 8.60 & 189 \\
     & Asian & 3.94 & 87 \\
    \textbf{Education Years} & Less than 16 & 54.56 & 1200 \\
     & More than 16 & 45.44 & 1000 \\
    \textbf{Diagnosis} & Cognitively Normal & 49.12 & 1081 \\
     & Mild Cognitive Impairment & 34.76 & 765 \\
     & Alzheimer’s Disease & 16.01 & 352 \\
    \textbf{APOE4 Status} & Heterozygous & 32.16 & 707 \\
     & Homozygous & 7.31 & 161 \\
     & None & 60.54 & 1332 \\
    \bottomrule
  \end{tabular}
\end{table}

\section{Predictors}
\label{sup:predictors}

\subsection{Architectural Design and Training}
\label{subsec:Architectural_Design_and_Training}
Our input data consists of multivariate features, including volumetric imaging features, demographic information, and clinical variables. All our predictors generate biomarker trajectories by using a learned model with an additional time input variable $t\in\setN_+$, that outputs an estimate of the corresponding future biomarker value $\Yhat_t$. Below we present five examples of such models, along with corresponding learning algorithms.

\textbf{Deep Kernel Regression.}
A fully connected feedforward neural network is used to linearly transform input data into a low-dimensional latent space. The transformed input is then passed to a Gaussian Process with a zero mean function and an radial basis function (RBF) kernel. The GP component is trained using exact inference by minimizing the negative marginal log-likelihood. 
For details on exact GP inference, refer to \citep{Rasmussen2004}. DKGP approach builds upon the deep kernel learning paradigm presented in \citep{wilson2016deep} and applied for trajectory prediction in \citep{pmlr-v193-tassopoulou22a}. The DKGP is trained for 100 epochs, using the Adam optimizer \citep{kingma2017adammethodstochasticoptimization} with weight decay 0.02 and a learning rate of 0.01 for the deep network parameters and 0.2 for the hyperparameters of the Gaussian Process. 

\textbf{Deep Mixed Effects.}
The DME model leverages an embedding network and a deep mean function to capture both global trends and local variations in structured regression tasks. The embedding network projects inputs into a 64-dimensional latent space, while the mean function, implemented as an MLP, maps the latent features to a scalar regression output. The Gaussian Process (GP) with an RBF kernel operates on the latent space to model residuals, with warping applied when an embedding function is used.
The model is trained end-to-end using variational inference, minimizing the negative Evidence Lower Bound (ELBO). Separate Adam optimizers are employed for the mean function and GP kernel parameters, each using a learning rate of \(10^{-3}\) and an \(L_2\)-penalty of \(10^{-3}\). For each training iteration, the GP parameters are adapted over \(n_{\text{adapt}} = 10\) steps with an inner learning rate of \(10^{-2}\). The training process spans 50 epochs, alternating between optimizing the mean function and GP parameters. Details on the DME model can be found in \citep{chung2019deepmixedeffectmodel}.

\textbf{Deep Quantile Regression.}
The model is a fully connected feedforward neural network designed for quantile regression, predicting multiple quantiles simultaneously (e.g., 0.1, 0.5, 0.9). It includes a 128-unit hidden layer and a 64-unit hidden layer, both followed by ReLU activations and dropout with a 0.2 rate for regularization. The output layer has one unit per quantile to estimate.
The model is trained using a quantile loss function \citep{koenker1978regression}. DQR is using Adam for the optimization, with a learning rate of 0.01 for 200 epochs. 

\textbf{Deep Regression with Monte Carlo Dropout.} The model is a fully connected feedforward neural network with Monte Carlo Dropout for uncertainty estimation in regression tasks. It consists of a 128-unit hidden layer, a 64-unit hidden layer, and a single-unit output layer for continuous scalar prediction. ReLU activations are applied to the hidden layers, and a dropout layer with a fixed rate 0.2 is applied during both training and inference to approximate Bayesian uncertainty. DRMC is using Adam for the optimization with learning rate of 0.01 for 200 epochs. 

\textbf{Bootstrap.} The model is a fully connected feedforward neural network for regression tasks, comprising a 128-unit hidden layer, a 64-unit hidden layer, and a single-unit output layer. ReLU activations are applied to hidden layers, and the output layer produces a raw scalar value for regression. The model is deployed as part of a bootstrap ensemble, where 10 instances are trained on different bootstrap-resampled subsets of the train data. Each instance is trained with consistent hyperparameters (epochs and learning rate) to enhance robustness and reduce variance by aggregating predictions.

\subsection{Predictive Standard Deviation}\label{subsec:Predictive_Standard_Deviation}

We outline the models and their uncertainty quantification mechanisms. Predictions  are denoted as \( \widehat{Y}_t \) with standard deviation \( \sigma(\Yhat_t) \).

\textbf{Deep Kernel Regression (DKGP) and Deep Mixed Effects (DME).}
Both models estimate the posterior predictive distribution using exact inference. The posterior predictive distribution is a gaussian distribution, and from that we extract the predictive mean that corresponds to the point estimates 
\( \widehat{Y}_t \) and the predictive variance that corresponds to a requested confidence level, i.e \(0.90\). From that we calculate the standard deviation \( \sigma(\Yhat_t) \). Details on the mathematical formula of the predictive mean and predictive variance can be found in \citep{Rasmussen2004}

\textbf{Deep Quantile Regression (DQR).}  
Deep Quantile Regression predicts the desired quantiles—lower, mean, and upper—for a specific confidence level. From these predictions, we calculate the quantiles corresponding to different percentiles, such as the 10th, 50th, and 90th percentiles, enabling uncertainty quantification. The confidence level represents the range between the lower and upper percentiles. The predicted variance is estimated using the spread between the upper and lower quantiles, adjusted by a z-score corresponding to the desired confidence level (e.g., 1.645 for a 90\% confidence interval). This calculation assumes that the predictive distribution is approximately Gaussian, allowing the z-score to be used as a scaling factor for the quantile spread. 

\textbf{Deep Regression with Monte Carlo (DRMC) dropout.}
Monte Carlo dropout approximates Bayesian inference via multiple stochastic forward passes with dropout. The model generates multiple predictions, and from these we extract the standard deviation of the sample. We run inference 100 times. 

\textbf{Bootstrap Deep Regression (Bootstrap).}
Bootstrap Deep Regression leverages an ensemble of 20 deep regression models, each trained on a different bootstrap sample, to estimate uncertainty. During inference, the ensemble produces multiple predictions for the same input, and the sample standard deviation of these predictions quantifies predictive uncertainty.

\section{Relation to the Broad Conformal Prediction Literature}
\label{subsec:conformal_literature}
Our conformal prediction method provides an extension to the setting of randomly timed trajectories, which is necessary for applying conformal prediction using real-world biomarker data. None of the previous methods, including \cite{Lin2022} can handle randomly-timed trajectories. Below, we elaborate on why prior CP methods for fixed-time trajectories cannot be applied in our setting of randomly timed biomarker trajectories. Our simple yet effective normalization function is able to accommodate conformal inference on randomly-timed trajectories. The normalizing function enables us to compute a normalized prediction error across all time steps, ensuring that no component within the maximum dominates the others in terms of scale. This function leverages the model’s predictive standard deviation to measure prediction uncertainty, effectively normalizing errors in a standard way. Our conformal prediction intervals at each time point in a test trajectory are not influenced by the trajectory’s length. The bound for the prediction error $|\hat{Y}_t - Y_t|$ at each time $t$ is derived by scaling the score $R$ by the time-specific uncertainty estimate $\sigma(\hat{Y}_t)$, yielding the value $R \cdot \sigma(\hat{Y}_t)$. This time-varying scaling ensures that the conformal bands adapt to the uncertainty at each time point. Additionally, a true ``worst-case" scenario would arise only if $\sigma(\hat{Y}_t) = 1$ at all time points, which does not hold here. 

Our extension of conformal prediction to randomly timed trajectories is necessary for real-world biomarker data. None of the previous methods \cite{Stankeviciute2021,Lin2022, Lindemann2023a,cleaveland2024time}can handle randomly-timed trajectories. Closest to our work are the approaches by \cite{cleaveland2024time} and \cite{Yu2023}, which are designed under the assumption of fixed-time trajectories---i.e., observations are made at the same pre-defined time points across the entire population. In contrast, our setting involves trajectories with observations collected at random times due to missed visits and variable scheduling typical in clinical studies. To adapt the method from \cite{cleaveland2024time, Yu2023} to our setting, one could select a subset of the data corresponding to a fixed set of time points (e.g., 3, 7, 9, and 12 months) and extract only the corresponding observations from each subject. However, this leads to substantial dataset reduction. In a simple experiment following this approach, we found that only 37 trajectories with observations at all four selected time points remained out of the total dataset of 2200 trajectories. Beyond dataset reduction, this method limits the applicability of the conformal prediction bands. More specifically, since \cite{cleaveland2024time} and \cite{Yu2023} define normalization factors $\sigma_t$ only at the pre-specified time points $t$ (e.g., 3, 7, 9, 12 months), their method can produce prediction intervals \textit{only} at those times during inference. Thus, if a test subject visits the clinic at months 4, 5, and 14, the available intervals at months 3, 7, 9, and 12, provided by \cite{cleaveland2024time} and \cite{Yu2023}, do not allow valid conformal inference at months 4, 5, and 14. In contrast, we introduce a normalizing function $\sigma(\cdot)$ that produces predictive uncertainty estimates $\sigma(\hat{Y}_t)$ for \textit{any} time point $t$, even if no observations at that time are available in the calibration data. This allows our method to produce valid conformal intervals at arbitrary test-time points, in contrast to existing conformal prediction methods designed for fixed-time trajectories. 

\section{Case Study: Details and Extended Results}
\label{sup:extended_results}

\subsection{Calibration Set Size Selection}\label{subsec:Calibration Set Size Selection}
In this section, we outline the procedure for determining the suitable calibration set size for conformal prediction. Specifically, we vary the fraction of the training set used for calibration from $0.01$ to $0.30$, while the remaining portion is reserved for predictor training. We then evaluate the resulting conformal intervals on a held-out validation set in terms of two key metrics: (i) mean coverage and (ii) mean interval width. 

This procedure is repeated for both hippocampus and ventricular volume biomarkers, at confidence levels of $0.90$, $0.95$, and $0.99$, and for all five conformal predictors. Our goal is to pick the size that achieves the desired coverage with the smallest possible interval width.

\begin{figure}[htb]
\vskip 0.1in
\begin{center}
    \centering
    \includegraphics[width=\linewidth]{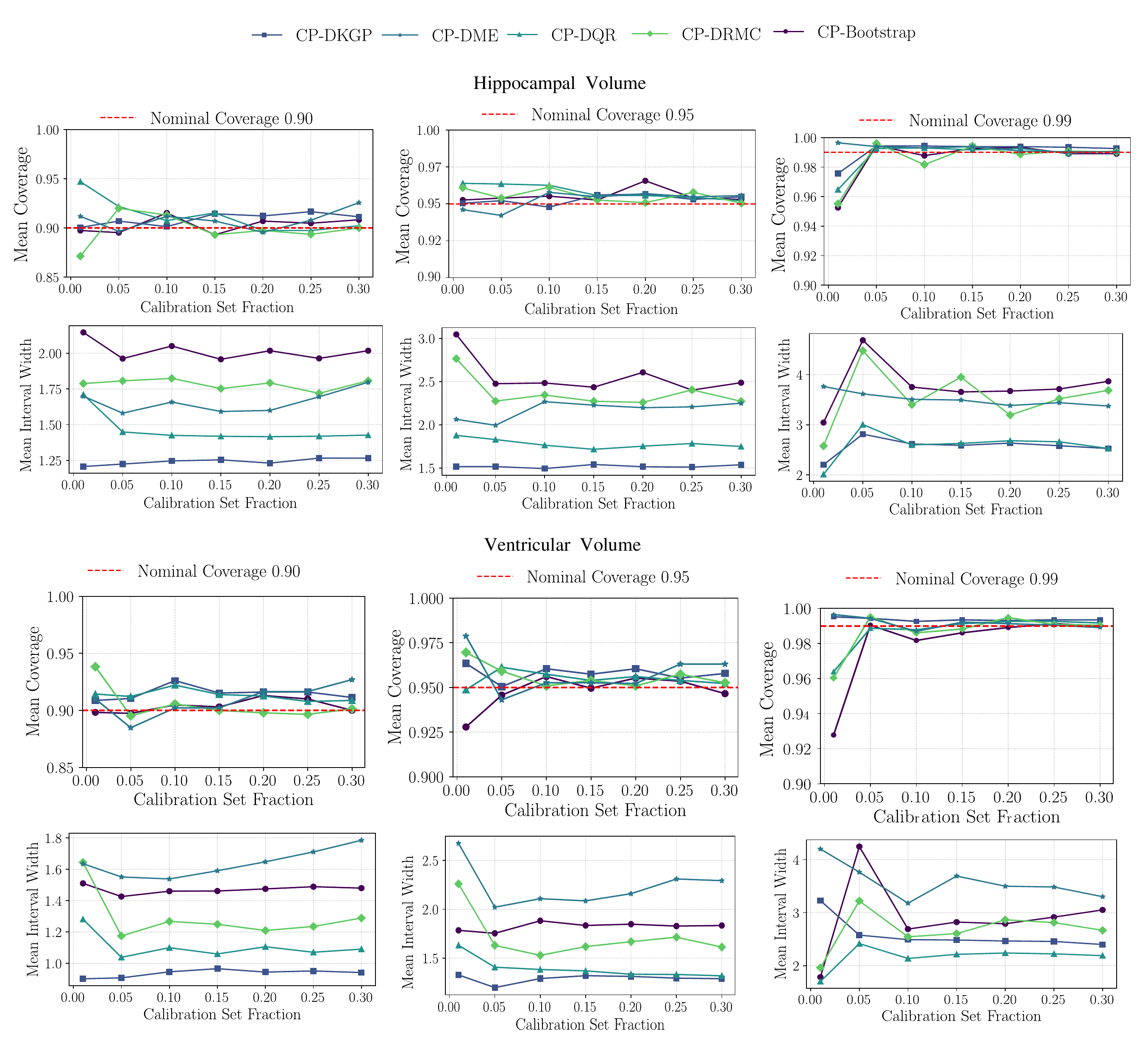}
    \caption{Mean coverage and mean interval width for hippocampal- and ventricular-volume trajectories at nominal coverage levels of \textit{0.90}, \textit{0.95}, and \textit{0.99}. Error bars denote the 95th percentile of the metrics across 10 data splits. Each column corresponds to a different nominal coverage level, and each curve represents one of five conformalized predictors (CP-DKGP, CP-DME, CP-DQR, CP-DRMC, CP-Bootstrap). Dashed red lines indicate the nominal coverage. The horizontal axis shows the fraction of the training set used for calibration, ranging from \textit{0.01} to \textit{0.30}}
    \label{fig:supplementary_calibration_set_size}
\end{center}
\vskip -0.1in
\end{figure}

Our findings, visualized in figure \ref{fig:supplementary_calibration_set_size} indicate that as the calibration set fraction increases, the empirical coverage exceeds the nominal coverage level. We identify a calibration set fraction of \textit{0.20} as a practical choice that achieves coverage while avoiding wide prediction intervals. This calibration set fraction corresponds to a calibration set with $414$ subjects.

\newpage
\subsection{Comparison between Baseline and Conformalized Predictors for Other Confidence Levels}\label{subsec:Comparison_with_Baselines_for_Other_Confidence Levels}
In this section, we provide additional results comparing the conformalized predictors with their baseline counterparts at confidence levels of $0.95$ and $0.99$. For both hippocampal- and ventricular-volume trajectories, we observe trends similar to those reported at the $0.90$ confidence level in Section~\ref{sec:Prediction_Brain_Biomarker} of the main paper. Specifically, conformal adjustments effectively boost empirical coverage to the desired nominal level. Gaussian Process based methods, such as the DKGP and DME already achieve the nominal coverage. Conformalized-DGKP and conformalized-DME attain nominal coverage with tighter bounds, which corresponds to less conservative uncertainty quantification. On the contrary, for the methods of DQR, DRMC and Boostrap, that provide undercoverage, their conformalized versions achive nominal coverage by widening the intervals.

These findings confirm the effectiveness of our conformal prediction across varying confidence levels.

\begin{figure}[ht]
\vskip 0.1in
\begin{center}
    \includegraphics[width=\linewidth]{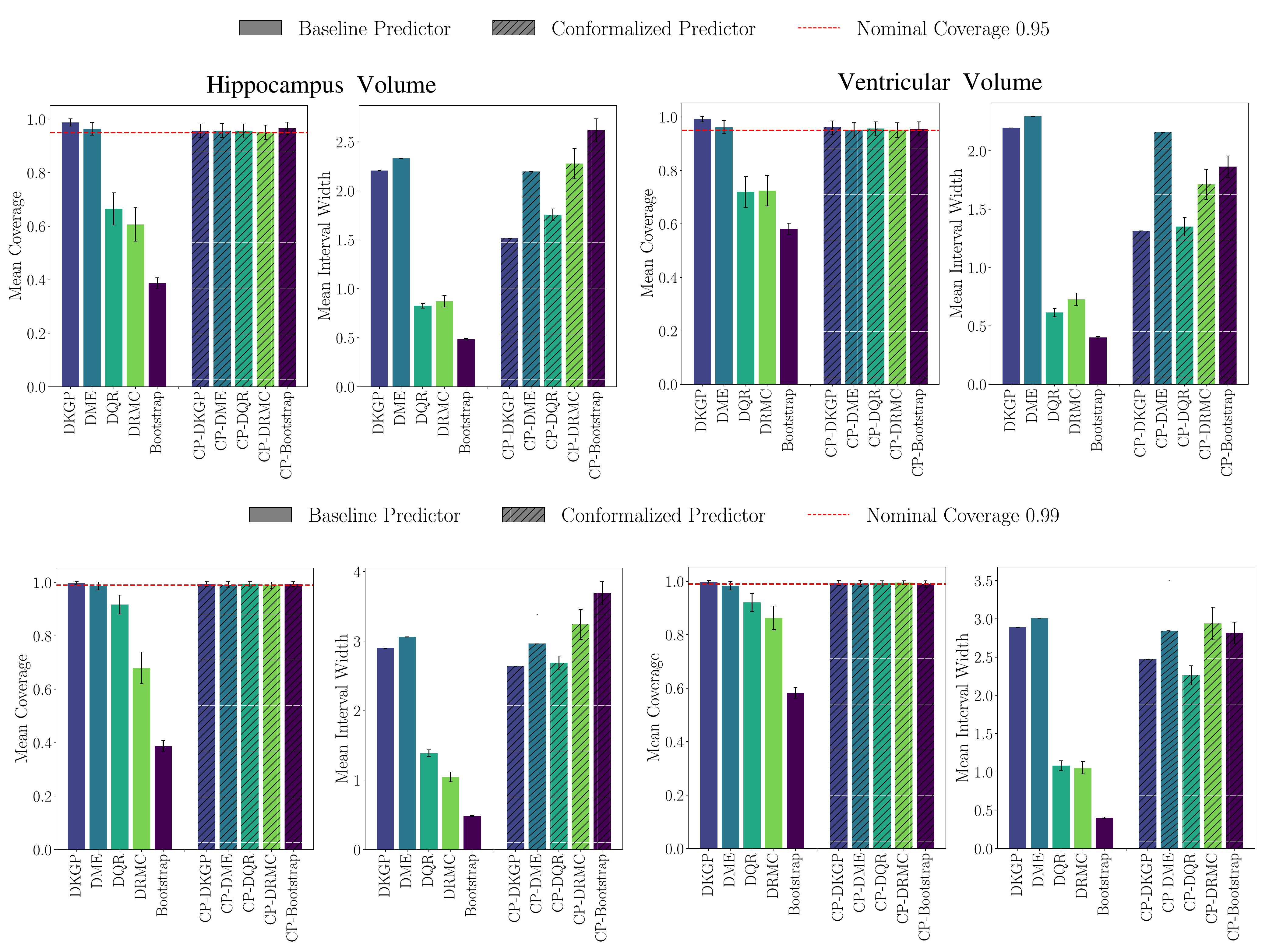}
    \caption{Mean coverage and mean interval width of baseline and conformalized predictors for hippocampal- and ventricular-volume trajectories at nominal confidence levels of 0.95 (top) and 0.99 (bottom). The dashed red line indicates the nominal coverage. Error bars denote the 95th percentile of the metrics across 10 data splits. Overall, the conformalized methods exhibit coverage closer to the nominal level, by adjusting accordingly the interval width.}
    \label{fig:allalphas}
\end{center}
\vskip -0.1in
\end{figure}

Next, we demonstrate how uncertainty changes over time starting from the initial acquisition, which, in our study, corresponds to the test subject's first hospital visit.

\subsection{Mean Interval Width of the Conformalized Predictors over Time}\label{subsec:Mean_Interval_Widths_over_Time}

Having established in the previous section that each conformalized baseline achieves desired coverage levels across different calibration set sizes and confidence levels, we now focus to the evolution of the conformal prediction intervals over time. In practical scenarios, the uncertainty in predicting future measurements naturally increases as we move further away from a known data acquisition, and understanding this growth in interval width is crucial for longitudinal analyses.

\begin{figure}[htb]
\vskip 0.1in
\begin{center}    
    \includegraphics[width=\linewidth]{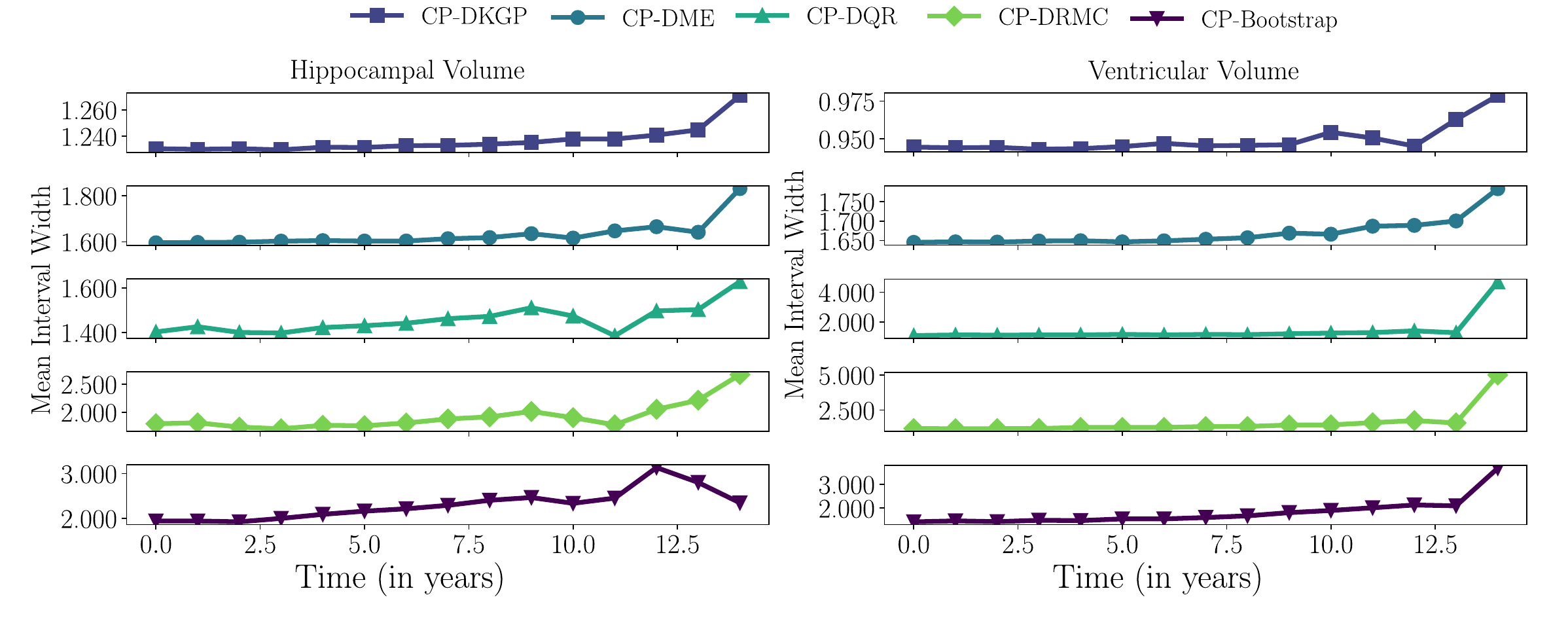}
    \caption{Temporal evolution of the mean interval width (averaged per year) for the five conformalized predictors. We observe that as time increases, the average conformal predictive intervals increase.}
    \label{fig:intervals_with_time_supp}
    \vskip -0.1in
\end{center}   
\end{figure}

Figure \ref{fig:intervals_with_time_supp} shows how the conformal prediction intervals of our selected methods vary with the time elapsed from the most recent (known) acquisition. Each row corresponds to a different conformalized predictor, while the horizontal axis indicates increasing time from the known data point, and the vertical axis shows the corresponding mean interval width. As expected, most methods exhibit relatively tight intervals when predicting only a short time ahead, but the intervals expand as the temporal distance grows. This behavior reflects the increasing uncertainty associated with predicting further into the future.

\newpage

\section{Group-Conditional Application across Covariate Subpopulations: Extended Results}
\label{sup:extended_stratif}

\begin{figure}[htb]
\vskip 0.1in
\begin{center}
    \includegraphics[width=\linewidth]{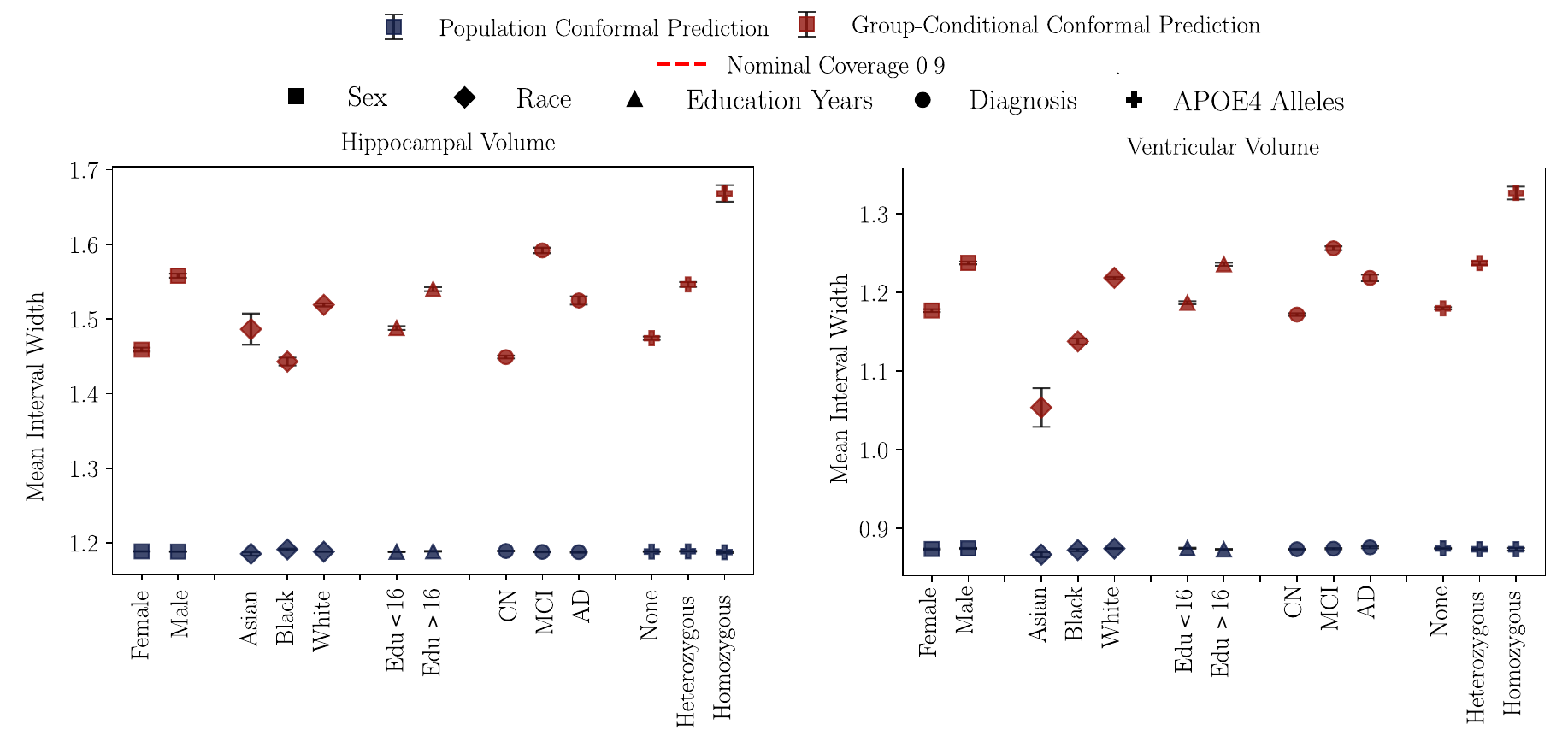}
    \caption{We compare the mean interval width of \textit{population} and \textit{group-conditional} conformal prediction bands for hippocampal- and ventricular-volume trajectories.Error bars denote the 95th percentile of the metrics across 10 data splits. Results are presented across five population stratifications based on individual covariates (sex, race, education level, cognitive diagnosis, and APOE4 alleles status). 
    We observe that \textit{group-conditional} intervals are adjusted per covariate group in order to reach the nominal coverage. On the contrary, the \textit{population} conformal prediction provides consistent intervals across all groups, that cause the coverage disparities.}
    \label{fig:intervals_stratif}
\end{center}
\vskip -0.1in
\end{figure}

Figure~\ref{fig:intervals_stratif} depicts the mean interval widths for hippocampal- and ventricular-volume trajectories across subgroups defined by demographic (Sex, Race, Education) and clinical (Diagnosis, APOE4 alleles) covariates. The population conformal prediction  generates narrower intervals for all the subgroups, leading to undercoverage, particularly for high-risk or underrepresented groups such as Black and Asian participants, MCI patients, and APOE4 homozygotes. In contrast, group-conditional conformalized predictors (in red) produces wider intervals in order to ensure that empirical coverage aligns with nominal levels within the specific subpopulations.
While for the hippoampal volume all the subpopulations were not covered, for the ventricular volume several subpopulations were already reaching the nominal coverage of 0.9, specifically for females, Asian, Black, subjects with less than 16 years of education as well as the Non-carriers. From the Figure ~\ref{fig:intervals_stratif} we observe that these are the ones that already reach the nominal level of coverage exhibit the lower increase in the interval width withing the subpopulation. For example, females have lower mean interval width than males. 

% \begin{figure}[!h]
% \vskip 0.1in
% \begin{center}
%     \includegraphics[width=0.5\linewidth]{images/main_paper/stratification_coverage_mainpaper_ventricles.pdf}
%     \caption{We compare the mean coverage of \textit{population} and \textit{group-conditional} conformal prediction bands for ventricular-volume trajectories. Error bars denote the 95th percentile of the metrics across 10 data splits. Results are presented across five population stratifications based on individual covariates (sex, race, education level, cognitive diagnosis, and APOE4 alleles status). 
%     We observe that \textit{group-conditional} intervals are adjusted per covariate group in order to reach the nominal coverage. On the contary, the \textit{population} conformal prediction provides consistent intervals across all groups, that cause the coverage disparities.}
%     \label{fig:intervals_stratif}
% \end{center}
% \vskip -0.1in
% \end{figure}

\begin{figure}[!ht]
\vskip 0.1in
\begin{center}
    \includegraphics[width=\linewidth]{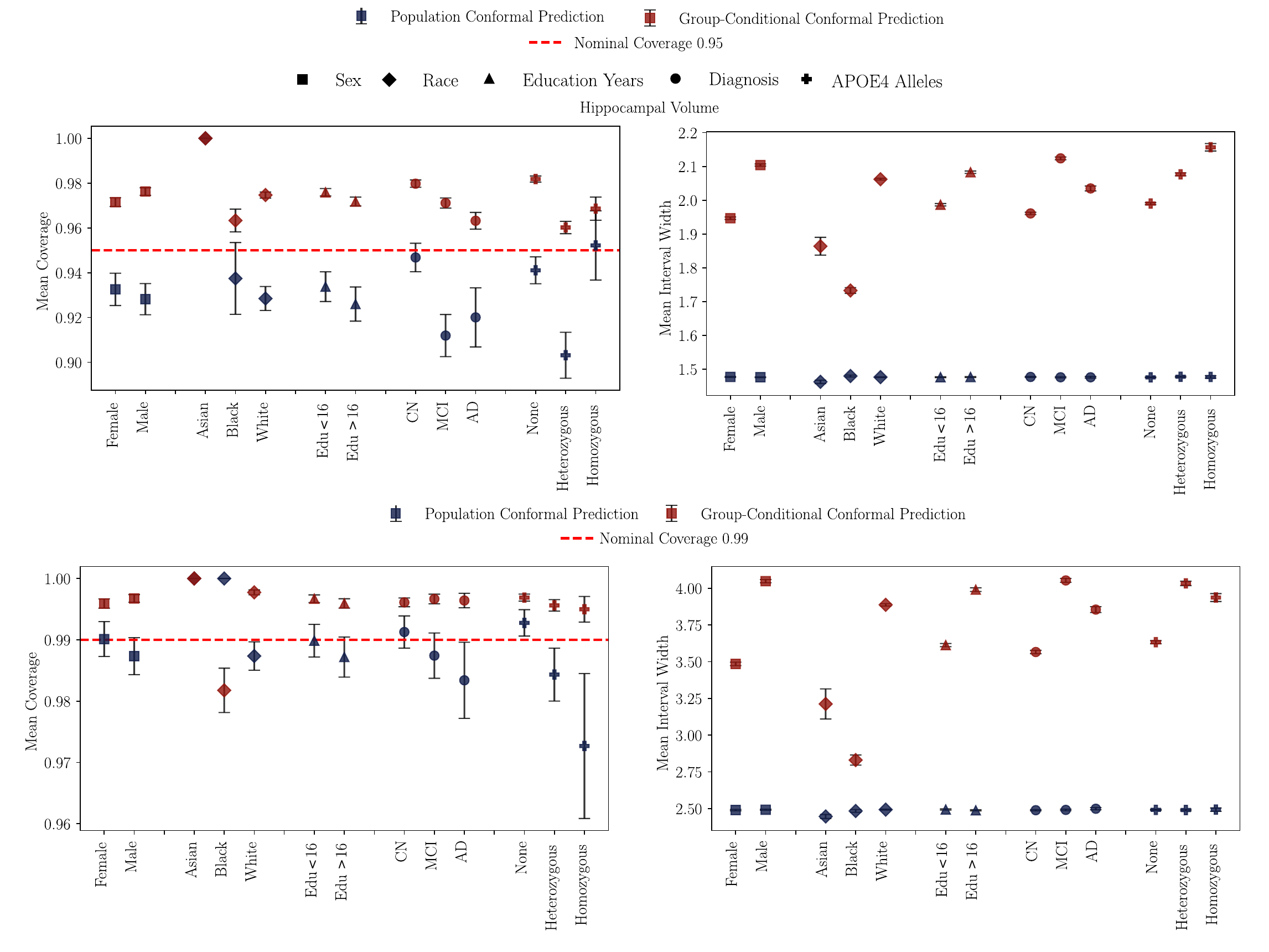}
    \caption{We compare the mean coverage of \textit{population} and \textit{group-conditional} conformal prediction bands for ventricular-volume trajectories. Error bars denote the 95th percentile of the metrics across 10 data splits. Results are presented across five population stratifications based on individual covariates (sex, race, education level, cognitive diagnosis, and APOE4 alleles status). 
    We observe that \textit{group-conditional} intervals are adjusted per covariate group in order to reach the nominal coverage. On the contrary, the \textit{population} conformal prediction provides consistent intervals across all groups, that cause the coverage disparities.}
    \label{fig:supp_hipp}
\end{center}
\vskip -0.1in
\end{figure}

\newpage

Figure~\ref{fig:supp_hipp} presents the empirical coverage and mean interval widths  for hippocampal-volume trajectories across subgroups at varying confidence levels $0.95$ and $0.99$. Again, we observe the same trend as in the Figure ~\ref{fig:stratified}, where population conformal prediction bands show noticeable undercoverage for underrepresented subgroups, such as Black and Asian participants, APOE4 homozygotes, and MCI patients. For the nominal level of $0.95$ only the Homozygotes appear to be covered by the population conformal predictor. Again, the group-conditional conformal prediction adapts the intervals of each subpopulation in order to achieve the nominal coverage withing subpopulation. For the nominal level of $0.99$ we observe that less subpopulations are miscovered and the majority of them align closely to the nominal level. This is expected as the confidence bands are wide enough in order to capture any subpopulation variability.   

\begin{figure*}[!ht]
\vskip 0.1in
\begin{center}
    \includegraphics[width=\linewidth]{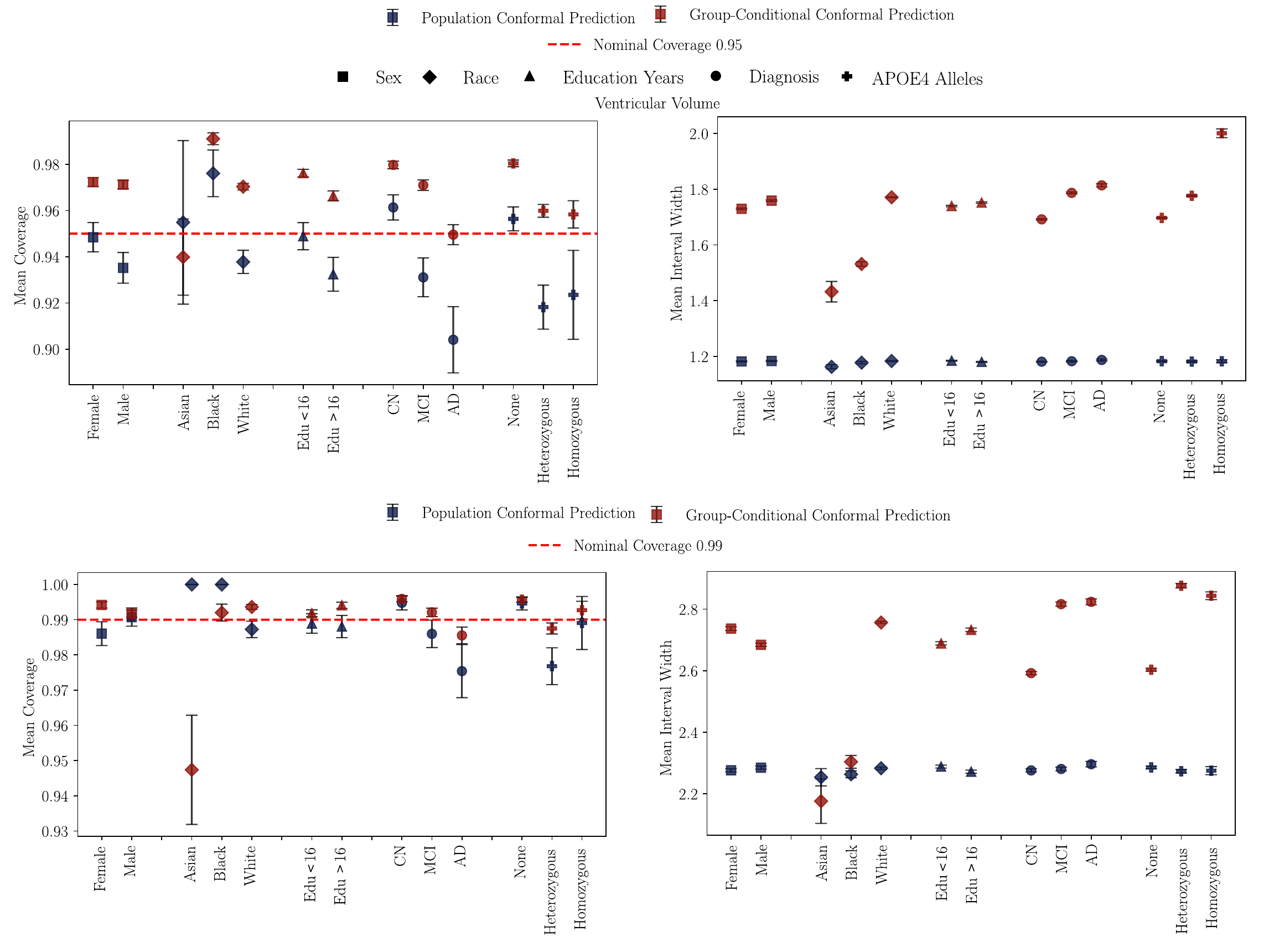}
    \caption{We compare the mean coverage and  mean interval width of \textit{population} and \textit{group-conditional} conformal prediction bands for  ventricular-volume trajectories for the nominal coverage of $0.95$ and $0.99$. Error bars denote the 95th percentile of the metrics across 10 data splits. Results are presented across five population stratifications based on individual covariates (sex, race, education level, cognitive diagnosis, and APOE4 alleles status). }
    \label{fig:coverage_and_intervals_plot_ventricle}
\end{center}
\vskip -0.1in
\end{figure*}

Figure~\ref{fig:coverage_and_intervals_plot_ventricle} presents the empirical coverage and mean interval widths for ventricular-volume trajectories across subgroups at varying confidence levels $0.95$ and $0.99$. For ventricular volume conformal prediction bands, a similar pattern is observed. Population conformal prediction underperform for the subgroups defined by Diagnosis and APOE4 status. Specifically, at the nominal level of 0.95, MCI and AD subjects as well as the heterozygotes and homozygotes are not covered by the population conformal prediction. However, the group-conditional conformal prediction tackles this by widening the intervals across those subgroups and thus achieving nominal coverage. 

\begin{figure*}[H]
\vskip 0.1in
\begin{center}
    \includegraphics[width=\linewidth]{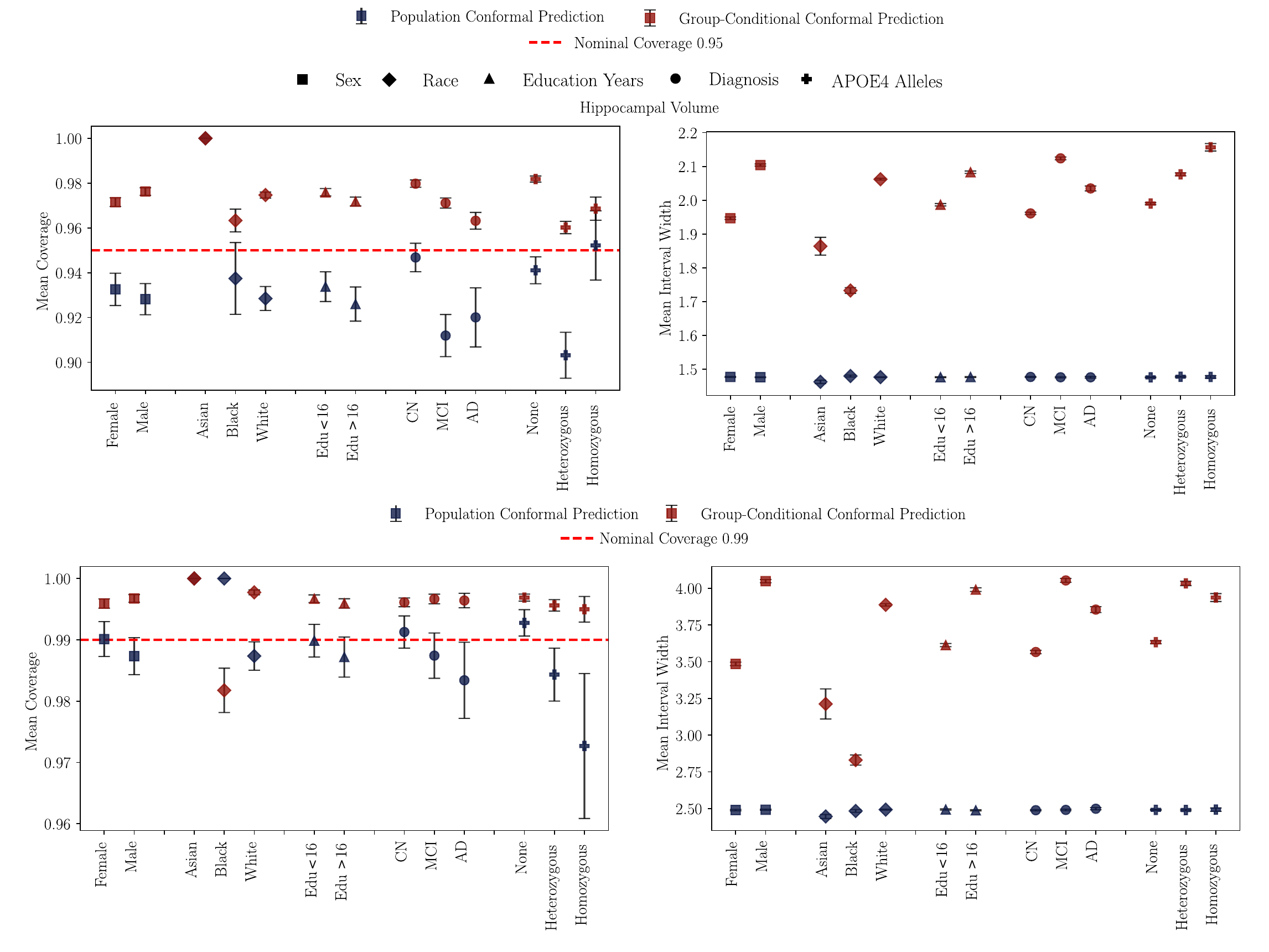}
    \caption{We compare the mean coverage and  mean interval width of \textit{population} and \textit{group-conditional} conformal prediction bands for hippocampal-volume trajectories at nominal confidence levels of $0.95$ and $0.99$. Error bars denote the 95th percentile of the metrics across 10 data splits. Results are presented across five population stratifications based on individual covariates (sex, race, education level, cognitive diagnosis, and APOE4 alleles status). }
    \label{fig:coverage_and_intervals_plot}
\end{center}
\vskip -0.1in
\end{figure*}

\newpage
\section{Clinical Application: Identification of High-Risk Alzheimer's Progressors}
\label{sup:clinical_applications}

To complement the clinical analysis in Section~\ref{sec:clinical_applications}, we further investigate the behavior of predictors that produce overconservative uncertainty estimates. Such models generate wide prediction intervals that capture most future outcomes, resulting in high recall but limited precision. While this behavior may be considered safe in clinical contexts, it can lead to over-inclusion of low-risk subjects and reduced specificity. Here, we examine whether conformalization can help refine these models by tightening the prediction bounds, thereby improving the balance between sensitivity and precision. We focus on two representative predictors: DME and DKGP. The total quantitative results are presented in Table \ref{tab:supp_mci_wcrc_full_ci}.

The DME model exhibits strongly conservative behavior. Using $\operatorname{\hat{RoC}}$, it identifies 81.4\% of MCI converters, and recall increases to 98.4\% under $\operatorname{RoCB}$, indicating near-complete inclusion of progressors. However, this recall comes at the cost of low precision—37.8\% for $\operatorname{\hat{RoC}}$ and 36.7\% for $\operatorname{RoCB}$—resulting in many false positives. Conformalization (CP--DME) slightly tightens the intervals, leading to a modest drop in $\operatorname{RoCB}$ recall to 94.2\%, while improving precision to 36.9\%.

The DKGP model also demonstrates overconservative behavior, producing wide prediction intervals that result in strong recall but imprecise discrimination. Under the $\operatorname{\hat{RoC}}$ metric, DKGP identifies 80.2\% of MCI converters, and $\operatorname{RoCB}$ pushes recall even higher to 92.2\%. However, this gain comes with reduced precision—falling from 50.6\% to 37.4\%. After conformalization, CP--DKGP produces slightly narrower intervals. While this modestly lowers recall under $\operatorname{\hat{RoC}}$ (to 72.1\%), the $\operatorname{RoCB}$-based recall remains high at 90.3\%, and precision increases slightly (to 37.6\%). These results highlight the role of conformalization in refining overly broad uncertainty estimates: by reducing excess conservativeness, CP--DKGP achieves a more balanced trade-off between identifying true progressors and limiting false positives, while still preserving the safety benefits of $\operatorname{RoCB}$-based risk stratification.

% \begin{table}[!h]
%   \centering
%   \caption{Youden-optimised discrimination on $z$-standardized rate of change (RoC) and lower-bound rate of change (LRoC) for MCI converters (full results).}
%   \label{tab:supp_mci_wcrc_full}
%   \begin{tabular}{lccccc}
%     \toprule
%     \textbf{Method} & \textbf{Metric} & $\boldsymbol{\tau^{\star}}$ & \textbf{Precision} & \textbf{Recall} & \textbf{F$_1$} \\
%     \midrule
%     DRMC       & RoC  & $-0.006$ & 0.436 & 0.671 & 0.528 \\
%                & LRoC & $-0.012$ & 0.403 & 0.884 & 0.553 \\
%     CP--DRMC   & RoC  & $-0.006$ & 0.432 & 0.740 & 0.546 \\
%                & LRoC & $-0.020$ & 0.395 & 0.915 & 0.552 \\
%     \midrule
%     DKGP       & RoC  & $-0.007$ & 0.506 & 0.802 & 0.621 \\
%                & LRoC & $-0.019$ & 0.374 & 0.922 & 0.532 \\
%     CP--DKGP   & RoC  & $-0.007$ & 0.507 & 0.721 & 0.595 \\
%                & LRoC & $-0.015$ & 0.376 & 0.903 & 0.531 \\
%     \midrule
%     DME        & RoC  & $-0.000$ & 0.378 & 0.814 & 0.516 \\
%                & LRoC & $-0.011$ & 0.367 & 0.984 & 0.535 \\
%     CP--DME    & RoC  & $-0.000$ & 0.378 & 0.814 & 0.516 \\
%                & LRoC & $-0.012$ & 0.370 & 0.942 & 0.531 \\
%     \midrule
%     Bootstrap  & RoC  & $-0.008$ & 0.501 & 0.698 & 0.583 \\
%                & LRoC & $-0.012$ & 0.407 & 0.837 & 0.548 \\
%     CP--Bootstrap & RoC  & $-0.007$ & 0.454 & 0.733 & 0.561 \\
%                   & LRoC & $-0.024$ & 0.387 & 0.888 & 0.539 \\
%     \bottomrule
%   \end{tabular}
% \end{table}

\begin{table}[!ht]
  \centering
  \caption{Youden-optimised discrimination on $z$-standardized rate of change (RoC) and lower rate-of-change bound ($\operatorname{RoCB}$) for MCI converters, with 95\% bootstrap confidence intervals (CIs).}
  \label{tab:supp_mci_wcrc_full_ci}
  \scriptsize
  \setlength{\tabcolsep}{4pt}
  \begin{tabular}{llcclll}
    \toprule
    \textbf{Method} & \textbf{Metric} & $\boldsymbol{\tau^{\star}}$ 
    & \textbf{Precision (95\% CI)} & \textbf{Recall (95\% CI)} & \textbf{F$_1$ (95\% CI)} \\
    \midrule
    DRMC       & $\operatorname{\hat{RoC}}$  & $-0.006$ & 0.436 [0.367, 0.455] & 0.671 [0.693, 0.919] & 0.528 [0.504, 0.593] \\
               & $\operatorname{RoCB}$ & $-0.012$ & 0.403 [0.367, 0.455] & 0.884 [0.693, 0.919] & 0.553 [0.504, 0.593] \\
    CP--DRMC   & $\operatorname{\hat{RoC}}$  & $-0.006$ & 0.432 [0.360, 0.446] & 0.740 [0.667, 0.956] & 0.546 [0.498, 0.590] \\
               & $\operatorname{RoCB}$ & $-0.020$ & 0.395 [0.360, 0.446] & 0.915 [0.667, 0.956] & 0.552 [0.498, 0.590] \\
    \midrule
    DKGP       & $\operatorname{\hat{RoC}}$  & $-0.007$ & 0.506 [0.340, 0.414] & 0.802 [0.807, 0.993] & 0.621 [0.494, 0.573] \\
               & $\operatorname{RoCB}$ & $-0.019$ & 0.374 [0.340, 0.414] & 0.922 [0.807, 0.993] & 0.532 [0.494, 0.573] \\
    CP--DKGP   & $\operatorname{\hat{RoC}}$  & $-0.007$ & 0.507 [0.338, 0.414] & 0.721 [0.809, 0.996] & 0.595 [0.491, 0.573] \\
               & $\operatorname{RoCB}$ & $-0.015$ & 0.376 [0.338, 0.414] & 0.903 [0.809, 0.996] & 0.531 [0.491, 0.573] \\
    \midrule
    DME        & $\operatorname{\hat{RoC}}$  & $-0.000$ & 0.378 [0.333, 0.406] & 0.814 [0.813, 1.000] & 0.516 [0.489, 0.571] \\
               & $\operatorname{RoCB}$ & $-0.011$ & 0.367 [0.333, 0.406] & 0.984 [0.813, 1.000] & 0.535 [0.489, 0.571] \\
    CP--DME    & $\operatorname{\hat{RoC}}$  & $-0.000$ & 0.378 [0.335, 0.408] & 0.814 [0.836, 1.000] & 0.516 [0.492, 0.569] \\
               & $\operatorname{RoCB}$ & $-0.012$ & 0.370 [0.335, 0.408] & 0.942 [0.836, 1.000] & 0.531 [0.492, 0.569] \\
    \midrule
    Bootstrap  & $\operatorname{\hat{RoC}}$  & $-0.008$ & 0.501 [0.363, 0.450] & 0.698 [0.754, 0.930] & 0.583 [0.503, 0.588] \\
               & $\operatorname{RoCB}$ & $-0.012$ & 0.407 [0.363, 0.450] & 0.837 [0.754, 0.930] & 0.548 [0.503, 0.588] \\
    CP--Bootstrap & $\operatorname{\hat{RoC}}$  & $-0.007$ & 0.454 [0.352, 0.432] & 0.733 [0.804, 0.956] & 0.561 [0.504, 0.581] \\
                  & $\operatorname{RoCB}$ & $-0.024$ & 0.387 [0.352, 0.432] & 0.888 [0.804, 0.956] & 0.539 [0.504, 0.581] \\
    \bottomrule
  \end{tabular}
\end{table}

\newpage

\subsection{\texorpdfstring{Threshold-Free Evaluation of $\operatorname{RoCB}$ vs $\operatorname{\hat{RoC}}$ Metrics}{Threshold-Free Evaluation of RoCB vs RoC Metrics}}
\label{sec:lroc_vs_roc}

To address concerns regarding the reliance on threshold-specific performance (e.g., Youden’s J), we conducted a comprehensive \textit{threshold-free} evaluation of our uncertainty-calibrated biomarker, $\operatorname{RoCB}$, compared against the predicted rate of change ($\operatorname{\hat{RoC}}$). This analysis spans ten trajectory prediction methods, including DKGP, CP-DRMC, Bootstrap, and others.

For each method, we computed ROC-AUC, PR-AUC, F1 score, precision, recall, and balanced accuracy across 720 test subjects. These metrics were chosen to reflect not only the discriminative ability (AUCs) but also the clinical decision trade-offs in high-risk detection (recall vs. precision).

$\operatorname{RoCB}$ consistently improves recall across all models, often by large margins (e.g., +62\% in CP-DQR). This demonstrates its strength in identifying high-risk individuals under worst-case prediction scenarios. However, this gain often comes at the cost of reduced precision and AUC metrics, reflecting a conservative, sensitivity-focused behavior. Despite this trade-off, F1-score improves in 6 out of 10 models, with the largest increase seen in CP-DQR (+9.8\%), indicating that in certain settings $\operatorname{RoCB}$ enhances both sensitivity and overall prediction balance.

These results confirm that $\operatorname{RoCB}$ is not a replacement for $\operatorname{\hat{RoC}}$, but a \textit{complementary metric} that prioritizes reliable early detection. This behavior is particularly beneficial in clinical contexts such as trial enrichment or preclinical screening, where high recall is often more critical than specificity.

\begin{table}[ht]
\centering
\caption{Threshold-free evaluation of $\operatorname{\hat{RoC}}$ vs. $\operatorname{RoCB}$ using ROC-AUC, PR-AUC, Recall, and F1-score. Values shown as ($\operatorname{\hat{RoC}}$ / $\operatorname{RoCB}$). LRoC improves recall consistently and maintains or improves F1 in 6 of 10 models.}
\label{tab:lroc_compact}
\begin{tabular}{lcccc}
\toprule
\textbf{Model} & \textbf{AUC} & \textbf{PR} & \textbf{Rec} & \textbf{F1} \\
\midrule
DRMC           & 0.597 / 0.541 & 0.421 / 0.354 & 0.671 / 0.884 & 0.528 / 0.553 \\
CP-DRMC        & 0.608 / 0.526 & 0.421 / 0.346 & 0.740 / 0.915 & 0.546 / 0.552 \\
DKGP           & 0.716 / 0.435 & 0.523 / 0.302 & 0.802 / 0.922 & 0.621 / 0.532 \\
CP-DKGP        & 0.697 / 0.438 & 0.502 / 0.303 & 0.721 / 0.903 & 0.595 / 0.531 \\
DQR            & 0.637 / 0.508 & 0.441 / 0.334 & 0.791 / 0.841 & 0.583 / 0.539 \\
CP-DQR         & 0.594 / 0.483 & 0.431 / 0.325 & 0.519 / 0.841 & \textbf{0.482 / 0.529} \\
DME            & 0.495 / 0.412 & 0.370 / 0.293 & 0.814 / 0.984 & 0.516 / 0.535 \\
CP-DME         & 0.495 / 0.411 & 0.370 / 0.293 & 0.814 / 0.942 & 0.516 / 0.531 \\
Bootstrap      & 0.686 / 0.520 & 0.495 / 0.343 & 0.698 / 0.837 & 0.583 / 0.548 \\
CP-Bootstrap   & 0.624 / 0.464 & 0.432 / 0.314 & 0.733 / 0.888 & 0.561 / 0.539 \\
\bottomrule
\end{tabular}
\end{table}

\end{document}